\documentclass[conference]{IEEEtran}
\usepackage{times}

\usepackage[numbers]{natbib}
\usepackage{multicol}
\usepackage[bookmarks=true]{hyperref}
\hypersetup{
    colorlinks=true,
    linkcolor=blue,
    filecolor=magenta,      
    urlcolor=cyan,
}
\usepackage[T1]{fontenc}
\usepackage[english]{babel}
\usepackage{url}
\usepackage{amsfonts}
\usepackage{nicefrac}
\usepackage{microtype}
\usepackage{algorithm}
\usepackage[noend]{algorithmic}
\usepackage{wrapfig}
\usepackage{amsmath}
\usepackage{amssymb}
\usepackage{mathtools}
\usepackage{subcaption}
\usepackage{ragged2e}
\usepackage{booktabs}
\usepackage{tikz}
\usepackage{etoolbox}
\usepackage{xspace}
\usepackage{dsfont}
\usepackage{xpatch}
\usepackage{enumerate}
\usepackage{xstring}
\usepackage{bbm}
\usepackage{amsthm}
\usepackage{setspace}
\usepackage{tabularx}
\usepackage{graphicx}

\newcommand{\algo}{DVD\xspace}
\newcommand{\sth}{Sth Sth V2\xspace}

\newcommand{\T}{\mathcal{T}}
\newcommand{\D}{\mathcal{D}}

\newcommand{\ca}{\mathcal}
\definecolor{mygreen}{rgb}{0.1, 0.6, 0.1}

\newcommand{\M}{\mathcal{M}}
\renewcommand{\S}{\mathcal{S}}
\newcommand{\A}{\mathcal{A}}
\newcommand{\di}{d_{i}}
\newcommand{\E}{\mathbb{E}}
\newcommand{\rt}{\mathcal{R}_\theta}

\begin{document}

\title{Learning Generalizable Robotic Reward Functions from ``In-The-Wild'' Human Videos}

\author{Annie S. Chen, Suraj Nair, Chelsea Finn \\ Stanford University}

\maketitle

\begin{abstract}
We are motivated by the goal of generalist robots that can complete a wide range of tasks across many environments. Critical to this is the robot's 
ability to acquire some metric of task success or reward, which is necessary for reinforcement learning, planning, or knowing when to ask for help.
For a general-purpose robot operating in the real world,
this reward function must also be able to generalize broadly across environments, tasks, and objects, while depending only on on-board sensor observations (e.g. RGB images).
While deep learning on large and diverse datasets has shown promise as a path towards such generalization in computer vision and natural language, collecting high quality datasets of robotic interaction at scale remains an open challenge.
In contrast, ``in-the-wild'' videos of humans (e.g. YouTube) contain an extensive collection of people 
doing interesting tasks across a diverse range of settings.
In this work, we propose a simple approach, Domain-agnostic Video Discriminator (\algo), 
that learns multitask reward functions by training a discriminator to classify whether two videos are performing the same task, and can
generalize by virtue of
learning from a \emph{small amount of robot data} with a \emph{broad dataset of human videos}. 
We find
that by leveraging diverse human datasets, this reward function (a) can generalize zero shot to unseen environments, (b) generalize zero shot to unseen tasks, and (c) can be combined with visual model predictive control to solve robotic manipulation tasks on a real WidowX200 robot in an unseen environment from a single human demo.

\end{abstract}

\IEEEpeerreviewmaketitle

\section{Introduction}

Despite recent progress in robotic learning on tasks ranging from grasping \cite{kalashnikov2018qtopt} to in-hand manipulation \cite{openai2019learning}, the long-standing goal of the ``generalist robot'' that can complete many tasks across environments and objects has remained out of reach.
While there are numerous challenges to overcome in achieving this goal, one critical aspect of learning general purpose robotic policies is the ability to learn \emph{general purpose reward functions}. Such reward functions are necessary for the robot to determine its own proficiency at the specified task from its on-board sensor observations (e.g. RGB camera images).
Moreover, unless these reward functions can generalize across varying environments and tasks, an agent cannot hope to use them to learn generalizable multi-task policies. 

\begin{figure}%
    \centering
    \vspace{0.1cm}
    \includegraphics[width=0.99\linewidth]{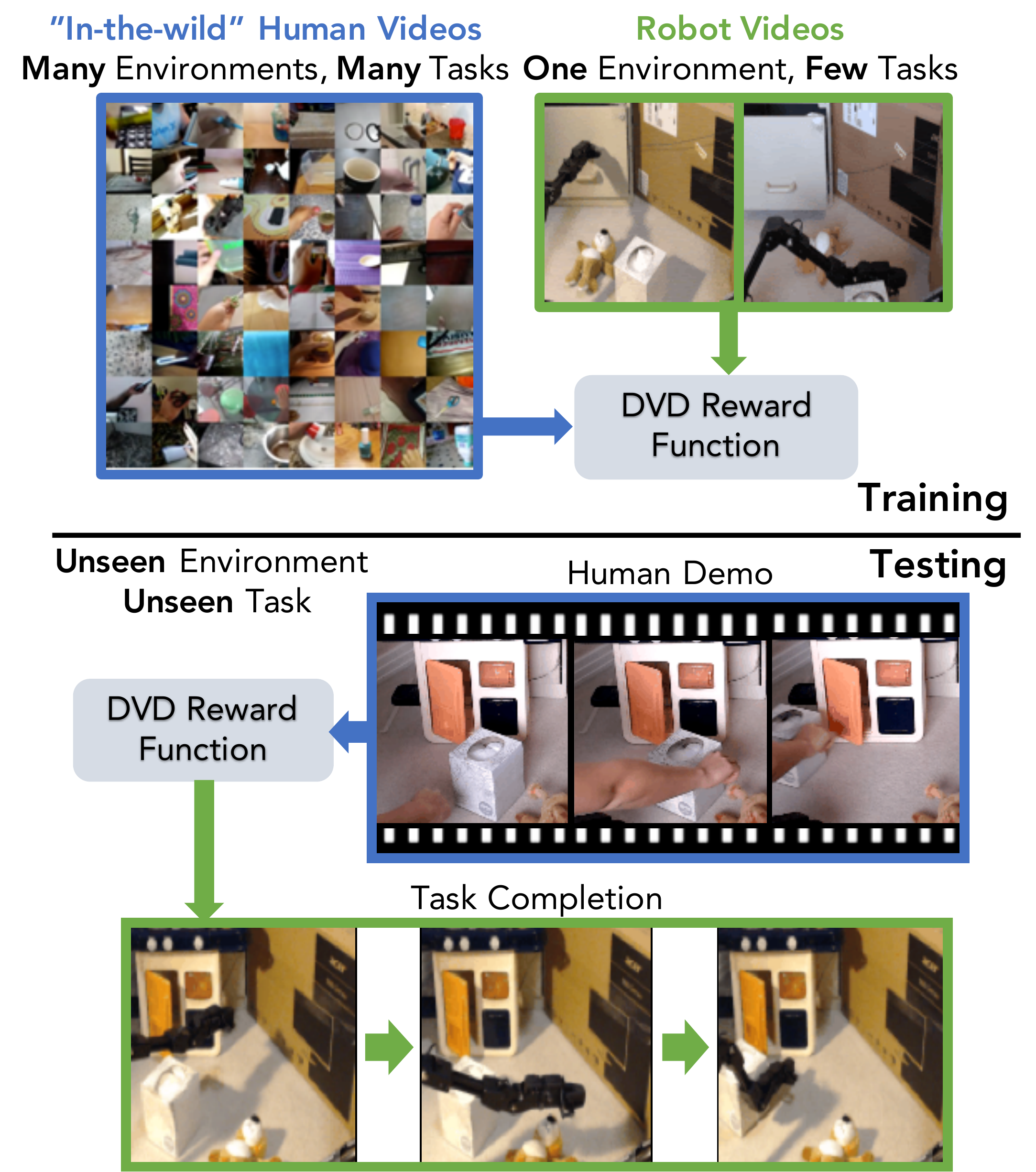}
    \caption{\small \textbf{Reward Learning and Planning from In-The-Wild Human Videos.} During training (\textbf{top}), the agent learns a reward function from a small set of robot videos in one environment, and a large set of in-the-wild human videos spanning many tasks and environments. At test time (\textbf{bottom}), the learned reward function is conditioned upon a task specification (a human video of the desired task), and produces a reward function which the robot can use to plan actions or learn a policy. By virtue of training on diverse human data, this reward function generalizes to unseen environments and tasks.}
    \vspace{-0.6cm}
    \label{pull_fig}
\end{figure}

While prior works in computer vision and NLP ~\cite{imagenet_cvpr09, bert, brown2020language} have shown notable generalization via large and diverse datasets, translating these successes to robotic learning has remained challenging, partially due to the dearth of broad, high-quality robotic interaction data. 
Motivated by this, a number of recent works have taken important steps towards the collection of large and diverse datasets of robotic interaction \cite{mandlekar2018roboturk, gupta2018robot, dasari2019robonet, young2020visual} 
and have shown some promise in enabling generalization \cite{dasari2019robonet}. 
At the same time, collecting such interaction data on real robots at a large scale remains challenging for a number of reasons, such as needing to balance data quality with scalability, and maintaining safety without relying heavily on human supervision and resets. Alternatively, YouTube and similar sources contain enormous amounts of ``in-the-wild'' visual data of humans interacting in diverse environments. Robots 
that can learn reward functions from such data have the potential to be able to generalize broadly due to the breadth of experience in this widely available data source.

Of course, using such ``in-the-wild'' data of humans for robotic learning comes with a myriad of challenges. First, such data often will have tremendous domain shift from the robot's observation space, in both the morphology of the agent and the visual appearance of the scene (e.g. see Figure~\ref{pull_fig}). Furthermore, the human's action space in these ``in-the-wild'' videos
is often
quite different from the robot's action space, and as a result there may not always be a clear mapping between human and robot behavior. Lastly, in practice these videos will often be low quality, noisy, and may have an extremely diverse set of viewpoints or backgrounds.
Critically however, this data is \emph{plentiful} and already exists, and is easily accessible through websites like YouTube or in pre-collected academic datasets like the Something-Something data set \cite{goyal2017something}, 
allowing them to be incorporated into the robot learning process with little additional supervision cost or collection overhead.

Given the above challenges, how might one actually learn reward functions from these videos? 
The key idea behind our approach is to train a classifier to predict whether two videos are completing the same task or not. By leveraging the activity labels that come with many human video datasets, along with a modest amount of robot demos, this model can capture the functional similarity between videos from drastically different visual domains. This approach, which we call a Domain-agnostic Video Discriminator (\algo), is simple and therefore can be readily scaled to large and diverse datasets, including heterogeneous datasets with both people and robots and without any dependence on a close one-to-one mapping between the robot and human data. Once trained, \algo~conditions on a human video as a demonstration, and the robot's behavior as the other video, and outputs a score which is an effective measure of task success or reward.

The core contribution of this work is a simple technique for learning multi-task reward functions from a mix of robot and in-the-wild human videos, which measures the functional similarity between the robot's behavior and that of a human demonstrator.
We find that this method is able to handle the diversity of human videos found in the Something-Something-V2 \cite{goyal2017something} dataset, and can be used in conjunction with visual model predictive control (VMPC) to solve tasks. Most notably,
we find that by training on diverse human videos (even from unrelated tasks), our learned reward function is able to more effectively generalize to unseen environments and unseen tasks than when only using robot data, yielding a 15-20\% absolute improvement in downstream task success. 
Lastly, we evaluate our method on a real WidowX200 robot, and find that it enables generalization to an unseen task in an unseen environment given only a single human demonstration video.

\section{Related Work}

\subsection{Reward Learning}
The problem of learning reward functions from demonstrations of tasks, also known as inverse reinforcement learning or inverse optimal control \cite{apprenticeship_abbeel}, has a rich literature of prior work
\cite{ratcliff_maxmargin, ziebart2008maximum, wulfmeier2016maximum, finn2016guided, fu2018learning}. A number of recent works have generalized this setting beyond full demonstrations to the case where humans provide only desired outcomes or goals \cite{fu2018variational, singh2019end}. Furthermore, both techniques have been shown to be effective for learning manipulation tasks on real robots in challenging high dimensional settings \cite{finn2016guided, singh2019end, Zhu2020The}.
Unlike these works,
which study single task problems in a single environment, the focus of this work is in learning generalizable \emph{multi-task} reward functions for visual robotic manipulation that can produce rewards for different tasks by conditioning on a single video of a human completing the task.

\subsection{Robotic Learning from Human Videos}

A number of works have studied learning robotic behavior from human videos. One approach is to explicitly perform some form of object or hand tracking in human videos, which can then be translated into a sequence of robot actions or motion primitives for task execution \cite{lee_2013_syntactic, yang_robot_WWW, nguyen2017translating, lee2017learning, rothfuss2018deep}. Unlike these works, which hand-design the mapping from a human sequence to robot behaviors, we aim to learn the functional similarity between human and robot videos through data.

More recently, a range of techniques have been proposed for end-to-end learning from human videos. One such approach is to learn to translate human demos or goals to the robot perspective directly through pixel based translation with paired \cite{liu2018imitation, sharma2019thirdperson} or unpaired \cite{smith2020avid} data. Other works attempt to infer actions, rewards, or state-values of human videos and use them for learning predictive models \cite{schmeckpeper2019learning} or RL \cite{edwards2019perceptual,schmeckpeper2020reinforcement}. Learning keypoint \cite{xiong2021learning, das2021modelbased} or object/task centric representations from videos \cite{sermanet2018timecontrastive, scalise2019improving, pirk2019online} is another promising strategy to learning rewards and representations between domains. Simulation has also been leveraged as supervision to learn such representations \cite{petrik2020learning} or to produce human data with domain randomization \cite{bonardi2019learning}. Finally, meta-learning \cite{daml} and subtask discovery \cite{sermanet2017unsupervised, goo_2019_multistep} have also been explored as techniques for acquiring robot rewards or demos from human videos.
In contrast to the majority of these works, which usually study a small set of human videos in a similar domain as the robot, we explicitly focus on leveraging ``in-the-wild'' human videos, specifically large and diverse sets of crowd-sourced videos from the real world from an existing dataset, which contains many different individuals, viewpoints, backgrounds, objects, and tasks. 

Our approach is certainly not the first to study using such in-the-wild human videos. Works that have used object trackers \cite{yang_robot_WWW}, simulation \cite{petrik2020learning}, and sub-task discovery \cite{goo_2019_multistep} have also been applied on in-the-wild video datasets like YouCook \cite{youcook}, Something-Something \cite{goyal2017something}, and ActivityNet \cite{caba2015activitynet}. Learning from such videos has also shown promise for navigation problems \cite{chang2020semantic}. Most related to this work is Concept2Robot \cite{shao2020concept}, which learns robotic reward functions using videos from the Something-Something dataset \cite{goyal2017something} by using a pretrained video classifier. Unlike Concept2Robot, our method learns a reward function that is conditioned on a human video demo, and thus can be used to generalize to new tasks. Furthermore, in Section~\ref{exp3}, we empirically find that our proposed approach provides a reward that generalizes to unseen environments with much greater success than the Concept2Robot classifier.

\begin{figure*}%
    \centering
    \vspace{0.1cm}
    \includegraphics[width=0.75\linewidth]{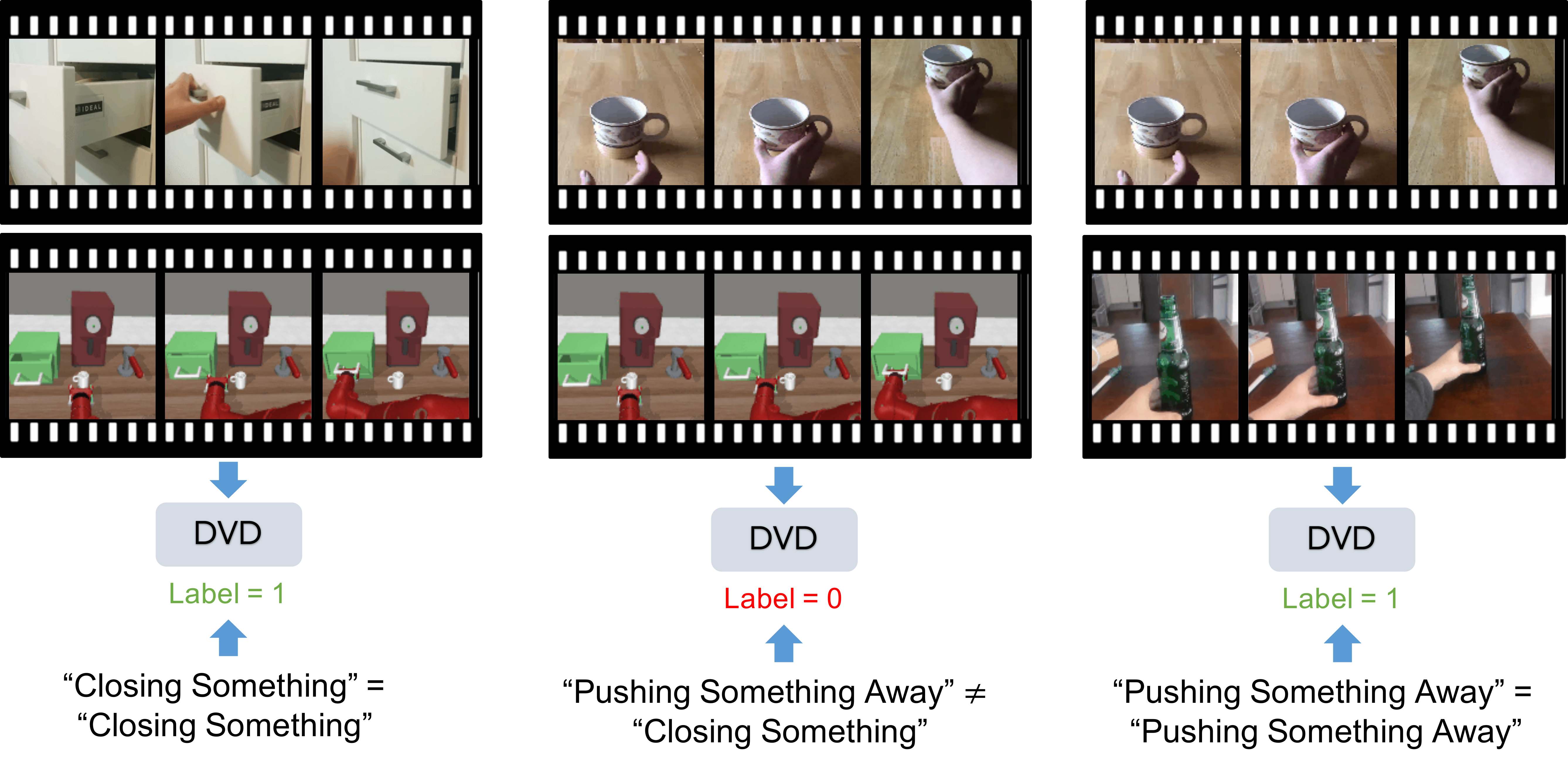}
    \vspace{-0.3cm}
    \caption{\small \textbf{Training \algo}. \algo is trained to predict if two videos are completing the same task or not. By leveraging task labels from in-the-wild human video datasets and a small number of robot demos, \algo is trained compare a video of a human to that of a robot (\textbf{left, middle}) and to compare pairs of human videos which may have significant visual differences, but may still be doing the same task (\textbf{right}). By training on these visually diverse examples, \algo is forced to learn the \emph{functional} similarity between the videos.}
    \label{classifier_fig}
    \vspace{-0.6cm}
\end{figure*}

\subsection{Robotic Learning from Large Datasets}

Much like our work, a number of prior works have studied how learning from broad datasets can enhance generalization in robot learning \cite{finn2017deep, pinto2016supersizing, zeng2018learning,  ebert2018visual, gupta2018robot, kalashnikov2018qtopt, dasari2019robonet, cabi2019scaling}. These works have largely studied the problem of collecting large and diverse robotic datasets in scalable ways \cite{mandlekar2018roboturk, gupta2018robot, dasari2019robonet, young2020visual, chen2020batch} as well as techniques for learning general purpose policies from this style of data in an offline \cite{ebert2018visual, cabi2019scaling} or online \cite{pinto2016supersizing, nair2018visual, kalashnikov2018qtopt} fashion. While our motivation of achieving generalization by learning from diverse data heavily overlaps with the above works, our approach fundamentally differs in that it aims to sidestep the challenges associated with collecting diverse robotic data by instead leveraging existing human data sources.

\section{Learning Generalizable Reward Functions with Domain-Agnostic Video Discriminators}
In this section, we describe our problem setting and introduce Domain-agnostic Video Discriminators~(\algo), a simple approach for learning reward functions that leverage in-the-wild human videos to generalize to unseen environments and tasks.

\subsection{Problem Statement}
In our problem setting, we consider a robot that aims to complete $K$ tasks $\{\T_i\}_{i=1}^{K}$, each of which has some underlying task reward function $\mathcal{R}_i$. As a result, for any given task $i$, our robotic agent operates in a fixed horizon Markov decision process (MDP) $\M^r_i$, consisting of the tuple $(\S, \A^r, p^r, \mathcal{R}_i, T)$ where $\S$ is the state space (in our case RGB images),
$\A^r$ is the robot's action space, $p^r(s_{t+1} | s_t, a^r_t)$ is the robot environment's stochastic dynamics, $\mathcal{R}_i$ indicates the reward for task $\T_i$, and $T$ is the episode horizon. Additionally, for each task $\T_i$, we consider a human operating in an MDP $\M^h_i$, consisting of the tuple $(\S, \A^h, p^h, \mathcal{R}_i, T)$ where $\A^h$ is the human's action space and $p^h(s_{t+1} | s_t, a^h_t)$ is the human environment's stochastic dynamics. 
Note that the human and robot MDPs for task $i$ share a state space $\S$, reward function $\mathcal{R}_i$, and horizon $T$, but may have different action spaces and transition dynamics.

We assume that the task reward functions $\mathcal{R}_i$ are unobserved, and need to be inferred through a video of the task. Note that for many tasks these rewards will not be Markovian--for example for the task of ``move two objects apart'', the reward depends not only on the current state, but on how close together the objects were initially. We instead assume that the reward at time $t$ is only dependent on the last $H < T$ timesteps, specifically states $s_{{t-H}:{t}}$.
Our goal then is to learn a parametric model which estimates the underlying reward function for each task, conditioned on a task-specifying video. That is, given (1) a sequence of $H$ states $s_{1:H}$ and (2) a video demonstration $\di = s^*_{1:t_{d_i}}$ of variable length for each task $\T_i$, we aim to learn a reward function $\rt(s_{1:{H}}, \di)$ that approximates $\mathcal{R}_i(s_{1:{H}})$ for each $i$. Such a non-Markovian reward can then be optimized using a number of strategies, ranging from open-loop planners to policies with memory or frame stacking.

For training the reward function $\rt$, we assume access to a dataset $\D^h = \{\D_{\T_i}^h\}_{i=1}^{N}$
of videos of humans doing $N < K$ tasks $\{\T_i\}_{i=1}^{N}$. There are no visual constraints on the viewpoints, backgrounds or quality of this dataset, and the dataset does not need to be balanced by task. We are also given a limited dataset $\D^r = \{\D_{\T_i}^r\}_{i=1}^{M}$ of videos of robot doing $M$ tasks $\{\T_i\}_{i=1}^M$ where $\{\T_i\}_{i=1}^M \subset \{\T_i\}_{i=1}^N$, and so $M \leq N$. Both datasets are partitioned by task. Since human data is widely available, we have many more human video demonstrations than robot video demonstrations per task and often many more tasks that have human videos but not robot videos, in which case $M << N$. 
Importantly, the reward is inferred only through visual observations and does not assume any access to actions or low dimensional states from either the human or robot data, and we do not make any assumptions on the visual similarity between the human and robot data. As a result, there can be a large domain shift between the two datasets. 

During evaluation, the robot is tasked with inferring the reward $\rt$ based on a new demo $\di$ specifying a task $\T_i$. The goal is for this reward to be effective for solving a task $\T_i$. Furthermore, we aim to learn $\rt$ in a way such that it can generalize to unseen tasks $\T_{new} \not\in \{\T_i\}_{i=1}^N$ given a task demonstration $d_{new}$.

\subsection{Domain-Agnostic Video Discriminators}

\begin{figure}%
    \centering
    \vspace{0.1cm}
    \includegraphics[width=0.99\linewidth]{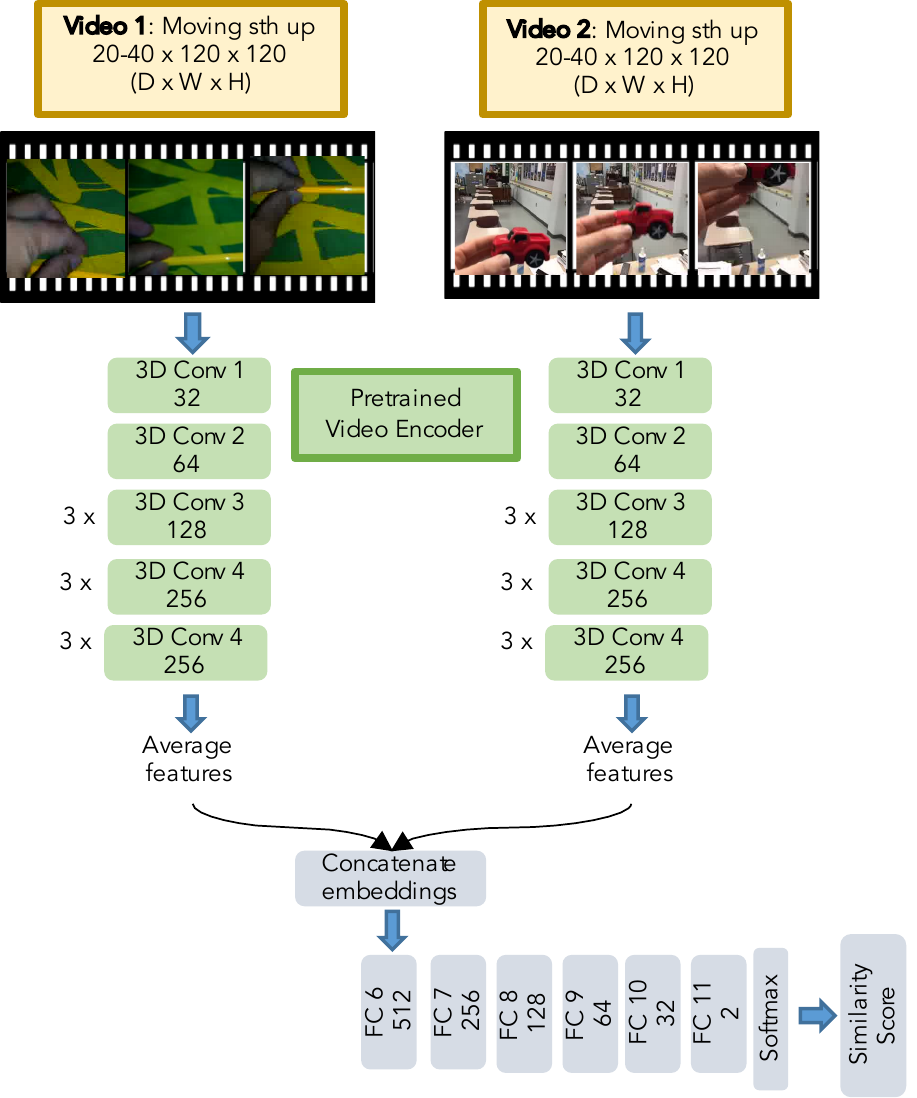}
    \vspace{-0.3cm}
    \caption{\small \textbf{\algo Architecture.} We use the same video encoder architecture as \cite{shao2020concept}. For each 3D convolution layer, the number of filters is denoted, and all kernels are $3 \times 3 \times 3$ except for the first, which is $3 \times 5 \times 5$. All conv layers have stride 1 in the temporal dimension, and conv layers 1, 3, 6, 9 and 11 have stride 2 in the spatial dimensions, the others having stride 1. All conv layers are followed by a BatchNorm3D layer and all layers except the last FC are followed by a ReLU activation.}
    \label{architecture}
    \vspace{-0.6cm}
\end{figure}

How exactly do we go about learning $\rt$? Our key idea is to learn $\rt$ that captures functional similarity
by training a classifier which takes as input two videos $\di$ from $\T_i$ and $d_j$ from $\T_j$ and predicts if $i=j$. Both videos can come from either $\D^h$ or $\D^r$, and labels can be acquired since we know which demos $d_i$ correspond to which tasks $\T_i$ (See Figure~\ref{classifier_fig}).

To train $\rt$, we sample batches of videos $(d_i, d_i', d_j)$
from $\D^h \cup \D^r$, where $d_i$ and $d_i'$ are both labelled as completing the same task $\T_i$, and $d_j$ is completing a different task $\T_j$. The output of $\rt$ represents a ``similarity score'' that indicates how similar task-wise the two input videos are. More formally, $\rt$ is trained to minimize the following objective, which is the average cross-entropy loss over video pairs in the distribution of the training data:
\begin{equation}
    \ca{J}(\theta) = \E_{\D^h \cup \D^r} [\log(\rt(d_i, d_i'))
    +   \log(1 - \rt(d_i, d_j))].
\label{eq:objective}
\end{equation}
Since in-the-wild human videos are so diverse and visually different from the robot environment, a large challenge lies in bridging the domain gap between the range of human video environments and the robot environment. In optimizing Equation~\ref{eq:objective}, $\rt$ must learn to identify functional behavior in the robot videos and associate it with actions in human videos.

\subsection{\algo Implementation}
\label{sec:method-DVD-implementation}

We implement our reward function $\rt$ as
\begin{equation}
\rt(d_i, d_j) = f_{sim}\left(f_{enc}(d_i), f_{enc}(d_j); \theta\right)
\label{eq:definition}
\end{equation}
where $h = f_{enc}$ is a pretrained video encoder and  $f_{sim}(h_i, h_j; \theta)$ is a fully connected neural network parametrized by $\theta$ trained to predict if video encodings $h_i$ and $h_j$ are completing the same task. Specifically, we encode each video using a neural network video encoder $f_{enc}$ into a latent space, and then train $f_{sim}$ as a binary classifier trained according to Equation~\ref{eq:objective}. See Figure~\ref{architecture} for the detailed architecture. 
$f_{enc}$ is pretrained on the entire \sth dataset and fixed during training (as in \cite{shao2020concept}), while $f_{sim}$ is randomly initialized.  
While the training dataset contains many more human videos than robot videos, we sample the batches so that they are roughly balanced between robot and human videos; specifically, each of $(d_i, d_i', d_j)$ are selected to be a robot demonstration with 0.5 probability.

\subsection{Using \algo~for Task Execution}
\label{sec:planning}

\begin{figure*}%
    \centering
    \vspace{0.1cm}
    \includegraphics[width=0.85\linewidth]{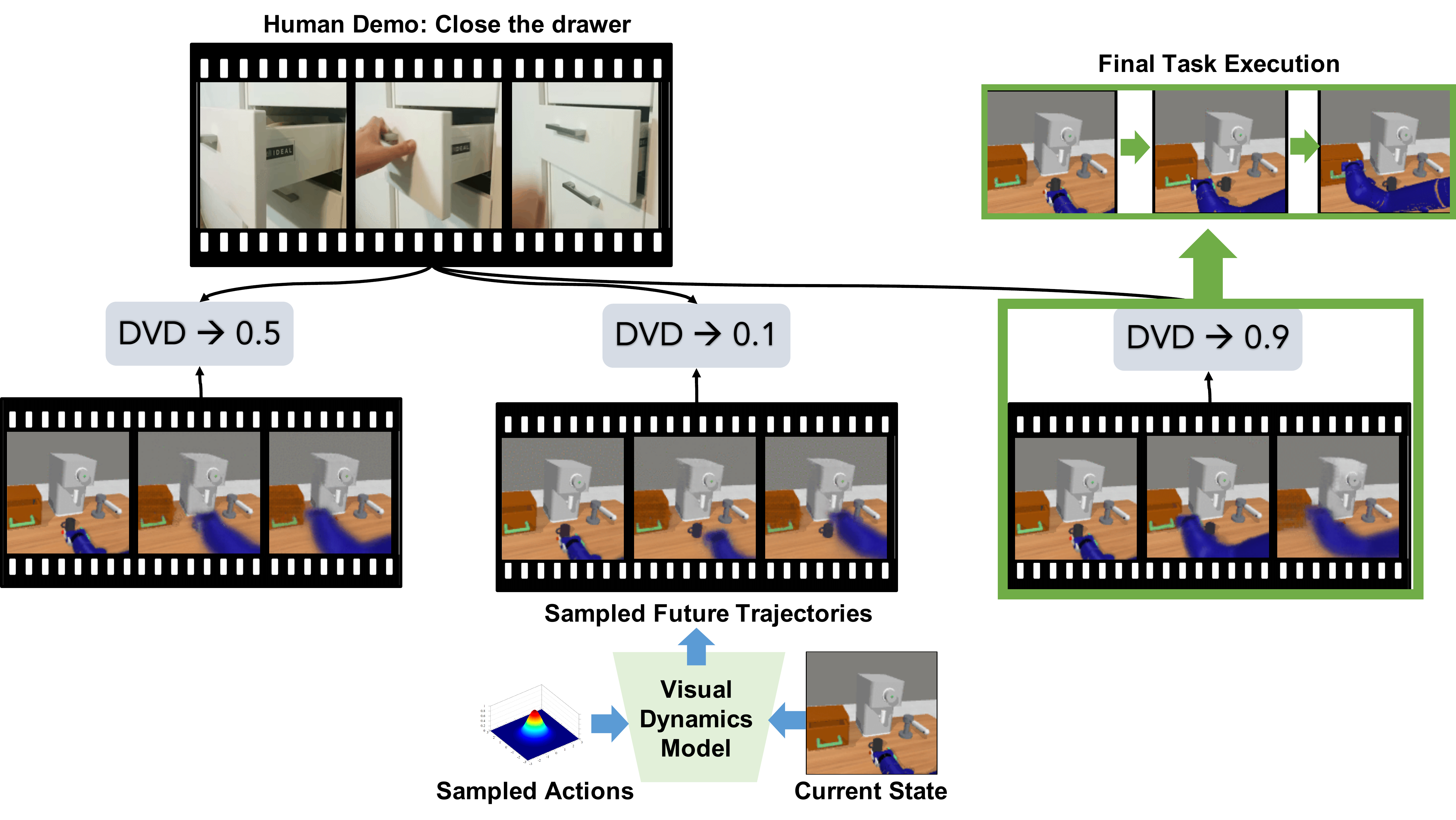}
    \vspace{-0.3cm}
    \caption{\small \textbf{Planning with \algo}. To use \algo to select actions, we perform visual model predictive control (VMPC) with a learned visual dynamics model. Specifically, we sample many action sequences from an action distribution and feed each through our visual dynamics model to get many ``imagined'' future trajectories. For each trajectory, we feed the predicted visual sequence into \algo along with the human provided demonstration video, which specifies the task. \algo scores each trajectory by its functional similarity to the human demo video, and steps the highest scored action sequence in the environment to complete the task.  }
    \label{planning_fig}
    \vspace{-0.3cm}
\end{figure*}

Once we've trained the reward function $R_\theta$, how do we use it to select actions that will successfully complete a task? 
While in principle, this reward function can be combined with either model-free or model based reinforcement learning approaches, we choose to use visual model predictive control (VMPC) \cite{watter2015embed, finn2017deep, ebert2018visual, hafner2019learning}, which uses a learned visual dynamics model to plan a sequence of actions. We condition $R_\theta$ on a human demonstration video $\di$ of the desired task $\T_i$ and then use the predicted similarity as a reward for optimizing actions with a learned visual dynamics model (See Figure~\ref{planning_fig}).

\begin{algorithm}[t]
\caption{\small \textsc{Domain-agnostic Video Discriminator (\algo)}}
\begin{algorithmic}[1]
\footnotesize
\STATE \textcolor{blue}{// Training \algo}
\STATE \textbf{Require:} $\D^h$ human demonstration data for $N$ tasks $\{\T_n\}$
\STATE \textbf{Require:} $\D^r$ robot demonstration data for $M$ tasks $\{\T_m\} \subseteq \{\T_n\}$
\STATE \textbf{Require:} Pre-trained video encoder $f_{enc}$
\STATE Randomly initialize $\theta$
\WHILE{training}
\STATE Sample anchor video $d_i \in \D^h \cup \D^r$
\STATE Sample positive video $d_i' \in  \{\D_{\T_i}^h\} \cup \{\D_{\T_i}^r\} \setminus {d_i} $
\STATE Sample negative video $d_j\in  \{\D_{\T_j}^h\} \cup \{\D_{\T_j}^r\} \forall j \neq i$ 
\STATE Update $\rt$  with $d_i, d_i', d_j$ according to Eq. \ref{eq:objective}
\ENDWHILE{}
\STATE \textcolor{blue}{// Planning Conditioned on Video Demo}
\STATE \textbf{Require:} Trained reward function  $\rt$
\& video prediction model $p_\phi$
\STATE \textbf{Require:} Human video demo $d_i$ for task $\T_i$
\FOR{trials $1, ..., n$}
    \STATE Sample $\{a^{1:G}_{1:H}\}$ \& get predictions $\{\tilde{s}^g_{1:H}\} \sim \{p_\phi(s_0, a^g_{1:H})\}$
    \STATE Step $a^*_{1:H}$ which maximizes  $\rt(\tilde{s}^g_{1:H}, d_i)$
\ENDFOR{}
\end{algorithmic}
\label{alg:training}
\end{algorithm}

Concretely, we first train an action-conditioned video prediction model $p_\phi(s_{t+1:t+H} | s_t, a_{t:t+H})$ using the SV2P model \cite{babaeizadeh2018stochastic}.
We then uses the cross-entropy method (CEM) \cite{rubinstein2013cross} with this dynamics model $p_\phi$ to choose actions that maximize similarity with the given demonstration. More specifically, for each iteration of CEM, for an input image $s_t$, we sample $G$ action trajectories of length $H$ and roll out $G$ corresponding predicted trajectories $\{s_{t+1:t+H}\}^g$ using $p_\phi$. We then feed each predicted trajectory and demonstration $\di$ into $\mathcal{R}_\theta$, resulting in $G$ similarity scores corresponding to the task-similarity between $\di$ and each predicted image trajectory. The action trajectory corresponding to the image sequence with the highest predicted probability
is then executed to complete the task.
The full algorithm with all stages is described in Algorithm~\ref{alg:training}.

\section{Experiments}
\label{sec:exps}
In our experiments, we aim to study how effectively our method \algo can leverage diverse human data, and to what extent doing so enables generalization to unseen environments and tasks.  
Concretely, we study the following questions: 
\begin{enumerate}
    \item By leveraging human videos is \algo able to more effectively \textbf{generalize to new environments?}
    
    \item By leveraging human videos is \algo able to more effectively \textbf{generalize to new tasks?}
    
    \item Does \algo enable robots to generalize from a single human demonstration \textbf{more effectively than prior work}?

    \item Can DVD infer rewards from a human video on a \textbf{real robot}?
\end{enumerate}

In the following sections, we first describe our experimental setup and then investigate the above questions. For videos please see \url{https://sites.google.com/view/dvd-human-videos}.

\subsection{Simulated Experimental Set-Up}

\begin{figure}%
    \centering
    \vspace{0.1cm}
    \includegraphics[width=0.99\linewidth]{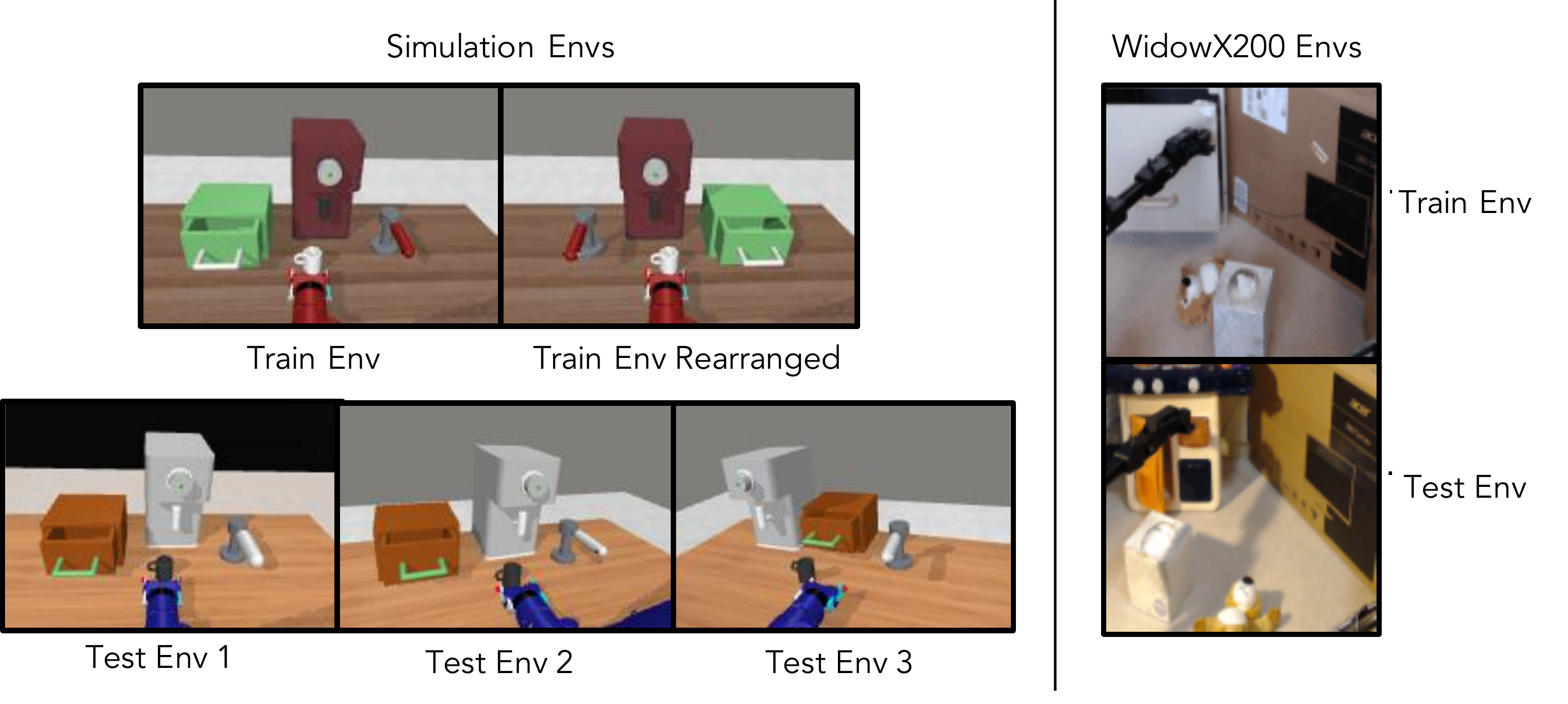}
    \vspace{-0.5cm}
    \caption{\small \textbf{Environment Domains.} We consider various simulated tabletop environments that have a drawer, a faucet, a coffee cup, and a coffee machine, as well as a real robot environment with a tissue box, stuffed animal, and either a file cabinet or a toy kitchen set. In the simulation experiments, half of the robot demonstrations that are used for training come from the train env and the other half from the rearranged train env.}
    \label{env_domains}
    \vspace{-0.6cm}
\end{figure}

\paragraph{Environments}
For our first 3 experimental questions, we utilize a MuJoCo \cite{mujoco} simulated tabletop environment adapted from Meta-World~\cite{yu2020meta} that consists of a Sawyer robot arm interacting with a drawer, a faucet, and a coffee cup/coffee machine.
We use 4 variants of this environment to study environment generalization, each of which is progressively more difficult, shown in Figure~\ref{env_domains}. These include an original variant (Train Env), from which we have task demos, as well as a variant with changed colors (Test Env 1), changed colors and viewpoint (Test Env 2), and changed colors, viewpoint, and object arrangement (Test Env 3). 

\begin{figure*} %
    \centering
    \vspace{0.1cm}
    \includegraphics[width=0.9\linewidth]{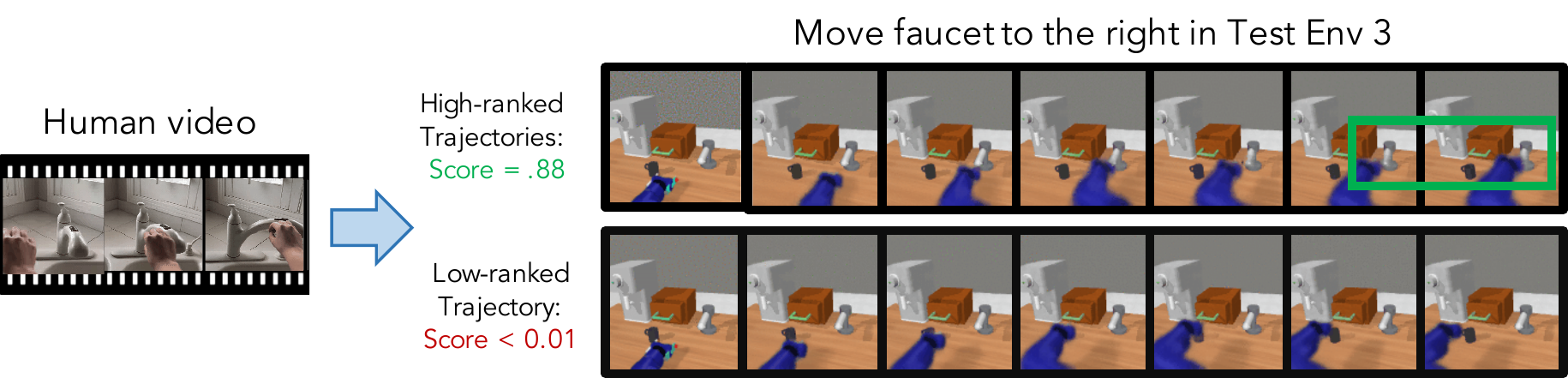}
    \caption{\small \textbf{Example Rankings During Planning.} Examples of predicted trajectories that are ranked high and low for the task of moving the faucet to the right in the test env 3 with the similarity scores that were outputted by \algo. \algo associates high functional similarity with trajectories that complete the same task as specified in the human video and low scores to trajectories that do not, despite the large visual domain shift between the given videos and the simulation environments.}
    \label{rankings_fig}
    \vspace{-0.3cm}
\end{figure*}

\paragraph{Tasks} We evaluate our method on three target tasks in simulation, specifically, (1) closing an open drawer, (2) turning the faucet right, and (3) pushing the cup away from the camera to the coffee machine. Each task is specified by an unseen in-the-wild human video completing the task (See Figure~\ref{rankings_fig}). 

\paragraph{Training Data}
For human demonstration data, we use the Something-Something-V2 dataset \cite{goyal2017something},
which contains 220,837 total videos and 174 total classes, each with humans performing a different basic action with a wide variety of different objects in various environments. 
Depending on the experiment, we choose videos from up to 15 different human tasks for training \algo, where each task has from 853-3170 videos (See Appendix for details).
For our simulated robot demonstration data, we assume 120 video demonstrations of 3 tasks \emph{in the training environment only} (See Figure~\ref{env_domains}). We ablate the number of robot demos needed in Section~\ref{sec:ablation}.

\subsection{Experiment 1: Environment Generalization}
\label{exp:env_gen}
In our first experiment, we aim to study how varying the amount of human data used for training impacts the reward function's ability to generalize across environments. To do so, we train \algo on robot videos of the 3 target tasks from the training environment, as well as varying amounts of human data, and measure task performance across \emph{unseen environments}.
One of our core hypotheses is that the use of diverse human data can improve the reward function's ability to generalize to new environments. To test this hypothesis, we compare training DVD on only the robot videos (\textbf{Robot Only}), to training DVD on a mix of the robot videos and human videos from $K$ tasks (\textbf{Robot + $K$ Human Tasks}). Note that the first 3 human tasks included are for the same 3 target tasks in the robot videos, and thus $K>3$ implies using human videos for \emph{completely unrelated} tasks to the target tasks. To evaluate the learned reward functions, we report the success rate from running visual MPC with respect to the inferred reward, where we train the visual dynamics model on data that is autonomously collected in the test environment. (See Appendix~\ref{sec:appendix-training} for details).
All methods infer the reward from a single human video. However, to provide an even stronger comparison, we also evaluate the Robot Only DVD model with a robot demo at test time \textbf{Robot Only (Robot Demo)}, since this model has only been trained with robot data.

\begin{figure}%
    \centering
    \vspace{0.1cm}
    \includegraphics[width=0.99\linewidth]{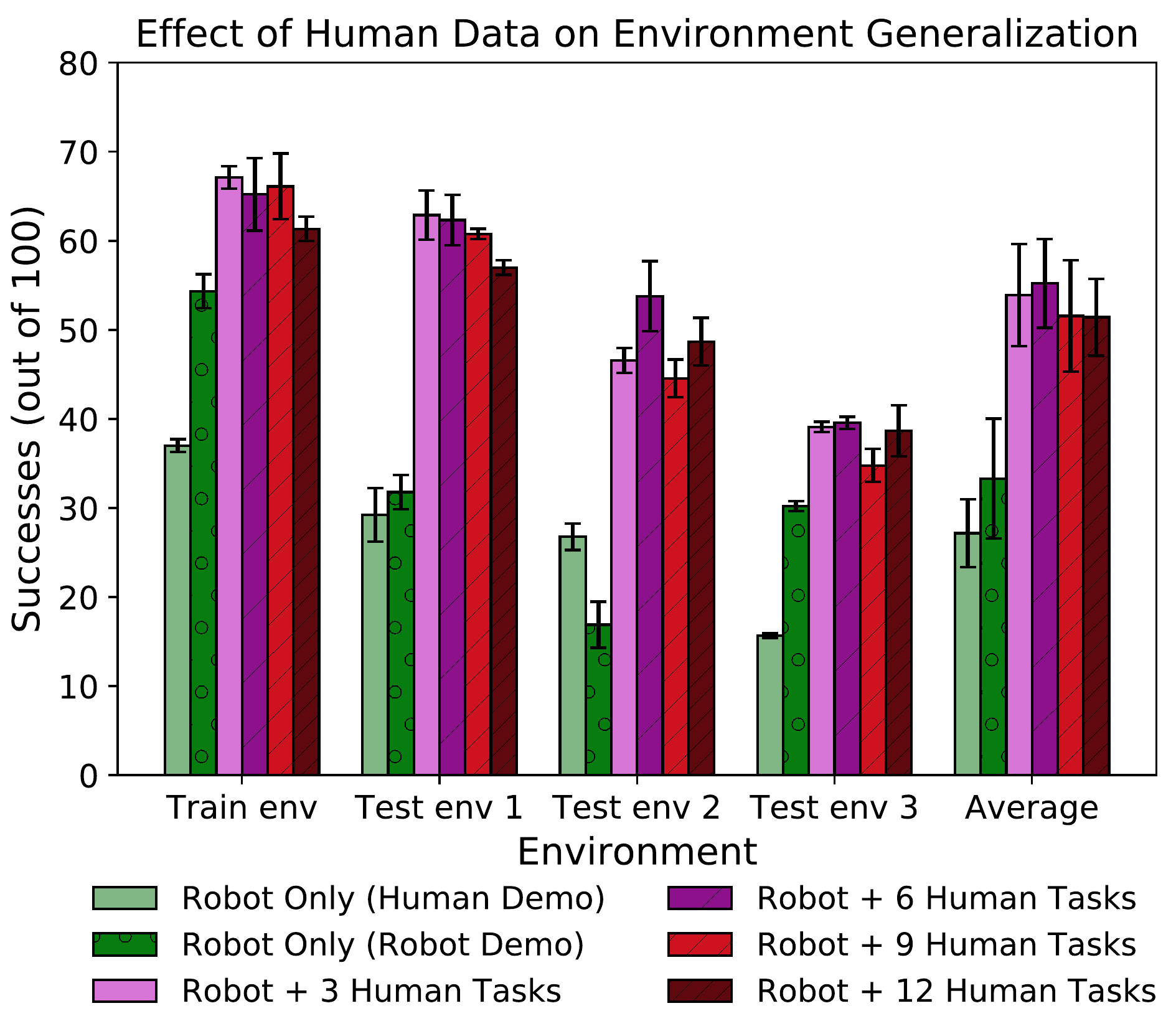}
    \vspace{-0.4cm}
    \caption{\small \textbf{Effect of Human Data on Environment Generalization.} We compare \algo's performance on seen and unseen environments when trained on only robot videos compared to varying number of human videos. We see that training with human videos provides significantly improved performance over only training on robot videos, and that \algo is generally robust to the number of different human video tasks used. Each bar shows the average success rate over all 3 target tasks, computed over 3 seeds of 100 trials, with error bars denoting standard error. }
    \vspace{-0.6cm}
    \label{env_gen}
\end{figure}

In Figure~\ref{env_gen}, we report the success rate using each reward function, computed over 3 randomized sets of 100 trials. Our \textit{first} key observation is that training with human videos significantly improves environment generalization performance over using only robot videos (20\% on average), even when the robot only comparison gets the privileged information of a robot demonstration. 
\textit{Additionally}, we observe that \algo~is generally robust to the number of human tasks included, even if these tasks are \emph{unrelated} to the target tasks. Even when using 9 completely unrelated tasks, performance greatly exceeds not using any human videos.
Qualitatively, in Figure~\ref{rankings_fig}, we observe that \algo gives high similarity scores to trajectories that are completing the task specified by the human video demo and low scores to trajectories that have less relevant behavior.

\subsection{Experiment 2: Task Generalization}
\label{exp:taskgen}
In our second experiment, we study how including human data for training affects the reward function's ability to generalize to new tasks. In this case, we do not train on any (human or robot) data from the target tasks, and instead train \algo on robot videos of the 3 \emph{different} tasks from the training environment, namely (1) opening the drawer, (2) moving something from right to left, (3) not moving any objects, as well as varying amounts of human data. 
To again test how human videos affect generalization, we compare the same methods as in the previous experiment.
Since we are testing task generalization, all evaluation is in the training environment.

\begin{table*}[t]
\centering
\center
\small
\begin{tabular}{ccccc}
\toprule
Method & Close drawer & Move faucet to right   & Push cup away from the camera  & Average                   \\
\cmidrule(lr){1-1} \cmidrule(lr){2-4} \cmidrule(lr){5-5}
Random     & 20.00 (3.00)  & 9.00 (1.73)  & 32.33 (8.08) & 20.44 (2.78) \\
Behavioral Cloning Policy        & 0.00 (0.00)  & 45.33 (38.84)  & 1.00 (0.00) & 15.44 (12.95) \\
Concept2Robot 174-way classifier        & n/a  & n/a  & n/a  & n/a  \\
\hline
\algo, Robot Only (Human Demo)  & \textbf{67.33 (4.51)} & 1.00 (1.00) & 29.67 (0.58) & 32.67 (1.53)  \\
\algo, Robot Only (Robot Demo) & 29.33 (14.99)  & 23.67 (1.53)  & 28.33 (0.58) & 27.11 (5.23) \\
\algo, Robot + 3 Human Tasks     & \textbf{66.33 (6.03)}  & 19.33 (0.58) & 40.00 (6.93)   & 41.89 (3.10) \\
\algo, Robot + 6 Human Tasks   & \underline{59.00 (5.29)}  & 17.00 (7.94) & \textbf{56.33 (11.06)} & 44.11 (1.39) \\
\algo, Robot + 9 Human Tasks     & 57.67 (0.58) & \textbf{52.67 (1.15)}  & \underline{55.00 (5.57)}  & \textbf{55.11 (2.04)} \\
\algo, Robot + 12 Human Tasks                & 31.67 (9.02) & \underline{49.00 (6.24)} & \textbf{57.33 (2.08)}  & \underline{46.00 (2.60)}  \\
\bottomrule
\end{tabular}
\vspace{-0.3cm}
\center
    \caption{\small
        Task generalization results in the original environment. \algo trained with human videos performs significantly better on average than with only robot videos, a baseline behavioral cloning policy, and random. We report the average success rate for all 3 target tasks, computed over 3 seeds of 100 trials, as well as the standard deviation in parentheses.
    }
\label{tab:task-gen}
\vspace{-0.3cm}
\end{table*}

In the bottom section of Table~\ref{tab:task-gen},
we report the success rate using \algo with varying amounts of human data, computed over 3 randomized sets of 100 trials. Similar to the conclusions of the environment generalization experiment, \textit{first} we find that training with human videos significantly improves task generalization performance over using only robot videos (by roughly 10\% on average), even with the robot only comparison conditioned on a robot demonstration. Given a human video demonstration, Robot Only does well at closing the drawer, but is completely unable to move the faucet to the right, suggesting that it is by default moving to the same area of the environment and is unable to actually distinguish tasks. This is unsurprising considering the reward function is not trained on any human videos. \textit{Second}, we observe that on average, including human videos for 6 unrelated human tasks can significantly improve performance, leading to more than a 20\% gap over just training with robot videos, suggesting that training with human videos from more unrelated tasks is particularly helpful for task generalization.

\subsection{Experiment 3: Prior Work Comparison}
\label{exp3}

In this experiment,  we study how effective \algo is compared to other techniques for learning from in-the-wild human videos. 
The most related work is \textbf{Concept2Robot} \cite{shao2020concept}, which uses a pretrained 174-way video classifier on only the \sth dataset (no robot videos) as a reward. Since this method is not naturally conducive to one-shot imitation from a video demonstration, during planning we follow the method used in the original paper and take the classification score for the target task from the predicted robot video as the reward (instead of conditioning on a human video). Unlike the open-loop trajectory generator used in the original paper, we use the same visual MPC approach for selecting actions for a fair comparison of the learned reward function; we expect the relative performance of the reward functions to be agnostic to this choice.
In addition, we also compare to a demo-conditioned \textbf{behavioral cloning} method,
similar to what has been used in prior work \cite{daml, bonardi2019learning, singh2020scalable}. 
We train this approach using behavior cloning on the 120 robot demonstrations and their actions for 3 tasks conditioned on a video demo of the task from either a robot or a human. See Appendix~\ref{sec:appendix-training} for more details on this comparison.
We also include a comparison to a \textbf{random} policy.

\begin{figure}%
    \centering
    \vspace{0.1cm}
    \includegraphics[width=0.99\linewidth]{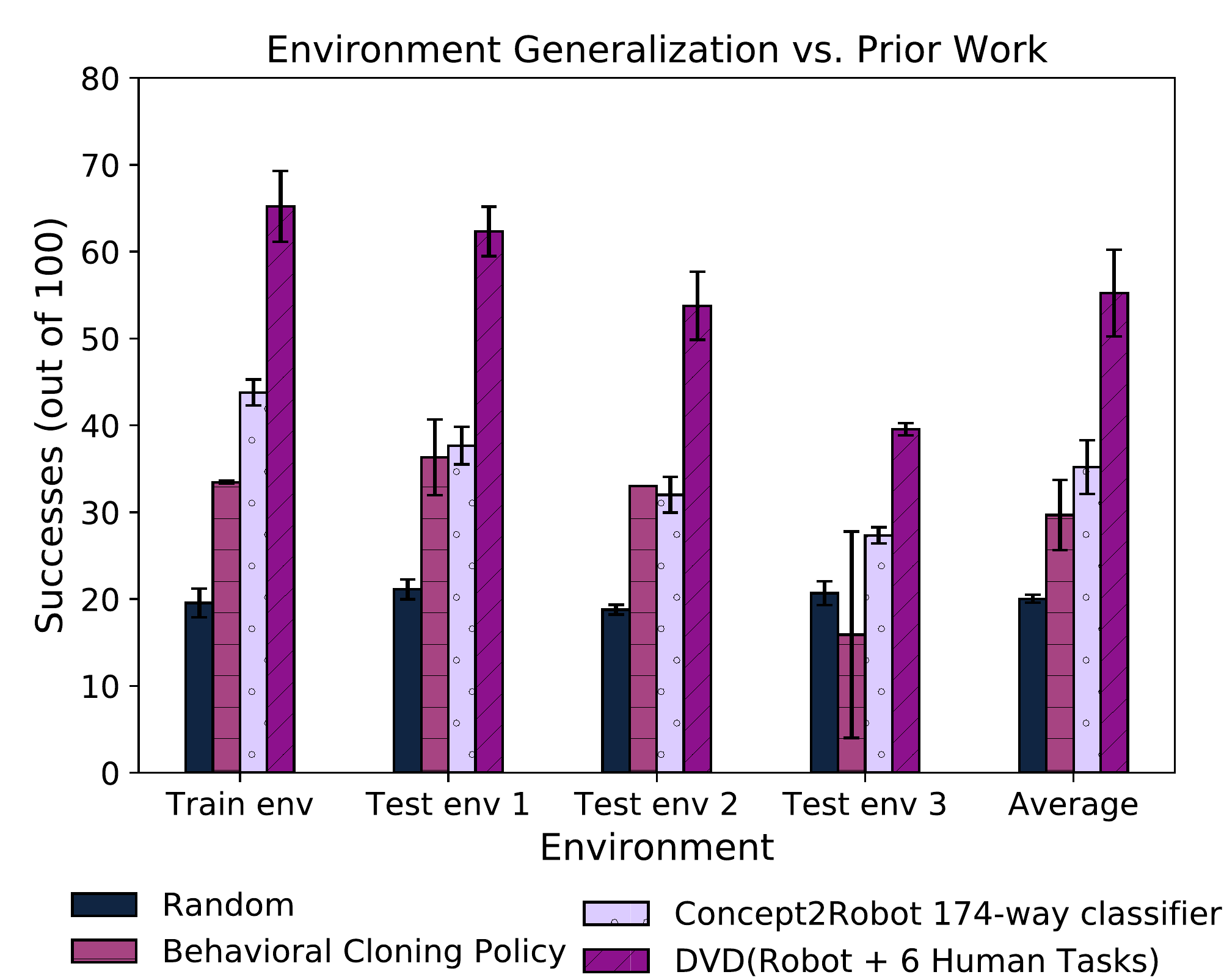}
    \caption{\small \textbf{Environment Generalization Prior Work Comparison.} Compared to Concept2Robot, the most relevant work leveraging ``in-the-wild'' human videos, as well as a demo-conditioned behavioral cloning policy and a random policy, \algo performs significantly better across all environments, and over 20\% better on average. Each bar shows the average success rate over all 3 target tasks, computed over 3 seeds of 100 trials, with error bars denoting standard error.}
    \vspace{-0.6cm}
    \label{pw}
\end{figure}

In Figure~\ref{pw} we compare \algo with 6 human videos to these prior methods on the environment generalization experiment presented in Section~\ref{exp:env_gen}. Across all environments, \algo performs significantly better than all three comparisons on the target tasks, and 20\% better on average than the best-performing other method. 
In Table~\ref{tab:task-gen}, we make the same comparison, now on the experiment of task generalization presented in Section~\ref{exp:taskgen}. Since Concept2Robot is not demo-conditioned and is already trained on all 174 possible human video tasks in the \sth dataset, there is no natural method for testing generalization to an unseen task specified by a human video. We see that \algo outperforms both other baselines by over 30\%.

\emph{First}, \algo's significant improvement over Concept2Robot suggests that in learning reward functions which address the human-domain robot gap, using \textit{some} robot data, even in small quantities, is important for good performance. \emph{Second}, methods like demo-conditioned behavior cloning likely require many more robot demonstrations to learn good policies, as prior work in demo-conditioned behavior cloning often use on the order of thousands of demonstrations \cite{singh2020scalable}. \algo~on the other hand, uses the demos only to learn a reward function and offloads the behavior learning to visual MPC. \emph{Lastly}, when examining the performance of demo-conditioned behavioral cloning on each individual task, we see the policy learns to ignore the conditioning demo and mimics one trajectory for one of the target tasks, doing well for only that task but completely failing at other tasks, suggesting that the policy struggles to infer the task from the visually diverse human videos.

\subsection{Experiment 4: Real Robot Efficacy}

To answer our last experimental question,
we study how \algo with human data enables better environment and task generalization on a real WidowX200 robot. We consider a similar setup as described in Sections~\ref{exp:env_gen} and~\ref{exp:taskgen}, where \algo~is now trained on 80 robot demos from each of 2 training tasks in a training environment and human videos. Then during testing, \algo is used as reward for visual MPC in an unseen environment, performing both a seen and unseen task. 

Specifically, in our real robot setup, the training environment consists of a file cabinet, and in the testing environment, it is replaced with a toy kitchen set (See Figure~\ref{env_domains}). The training tasks are ``Closing Something'' and ``Pushing something left to right'' and the test tasks are ``Closing Something'' (seen) and ``Pushing something right to left'' (unseen). 

We compare \algo~with varying amounts of human data to only robot data and baselines in Table~\ref{tab:robot}, where we report the success rate out of 20 trials when used with visual MPC conditioned on a human demo of the task. \algo trained with human videos has about twice the success rate
when leveraging the diverse human dataset than when relying only on robot videos. In particular, \algo trained with 6 tasks worth of human videos succeeds over 65-70\% of the time whereas robot only succeeds at most 40\%. We also observe that in general using human videos from unrelated tasks improves over only using human videos for the training tasks.
Finally, we see qualitatively in Figure~\ref{fig:realrobot_rankings} in Appendix~\ref{sec:appendix-exp-results} that \algo captures the functional task being specified, in this case closing the door.

\begin{table}[]
\centering
\center
\small
\begin{tabular}{ccc}
\toprule
Method (Out of 20 Trials) & Test Env &  \begin{tabular}[c]{@{}c@{}}Test Env + \\ Unseen Task \end{tabular} \\
\cmidrule(lr){1-1} \cmidrule(lr){2-2} \cmidrule(lr){3-3}
Random                           & 5 & 5  \\
Concept2Robot 174-way classifier & 4 & n/a \\
\hline
\algo, Robot Only (Human Demo)      & 5 & 6  \\
\algo, Robot Only (Robot Demo)      & 5 & 8  \\
\algo, Robot + 2 Human Tasks          & 7 & 7  \\
\algo, Robot + 6 Human Tasks          & \textbf{13} & \textbf{14} \\
\algo, Robot + 9 Human Tasks          & 9 & 11 \\
\algo, Robot + 12 Human Tasks         & 10 & 9  \\
\bottomrule
\end{tabular}
\vspace{-0.3cm}
\center
    \caption{\small
        \textbf{Env and task generalization results on a real robot.} We report successes out of 20 trials on a WidowX200 in an unseen environment on two different tasks, one on closing a toy kitchen door and another on moving a tissue box to the left. On both, \algo performs significantly better when trained with human videos than with only robot demonstrations.
    }
\label{tab:robot}
\vspace{-0.6cm}
\end{table}

\subsection{Ablation on Amount of Robot Data for Training}
\label{sec:ablation}

In our previous simulation experiments, we use 120 robot demonstrations per task. 
While this is a manageable number of robot demonstrations, it would be better to rely on fewer demonstrations. Hence, we ablate on the number of robot demonstrations used during training and evaluate environment generalization. 
\begin{figure}%
    \centering
    \vspace{0.1cm}
    \includegraphics[width=0.99\linewidth]{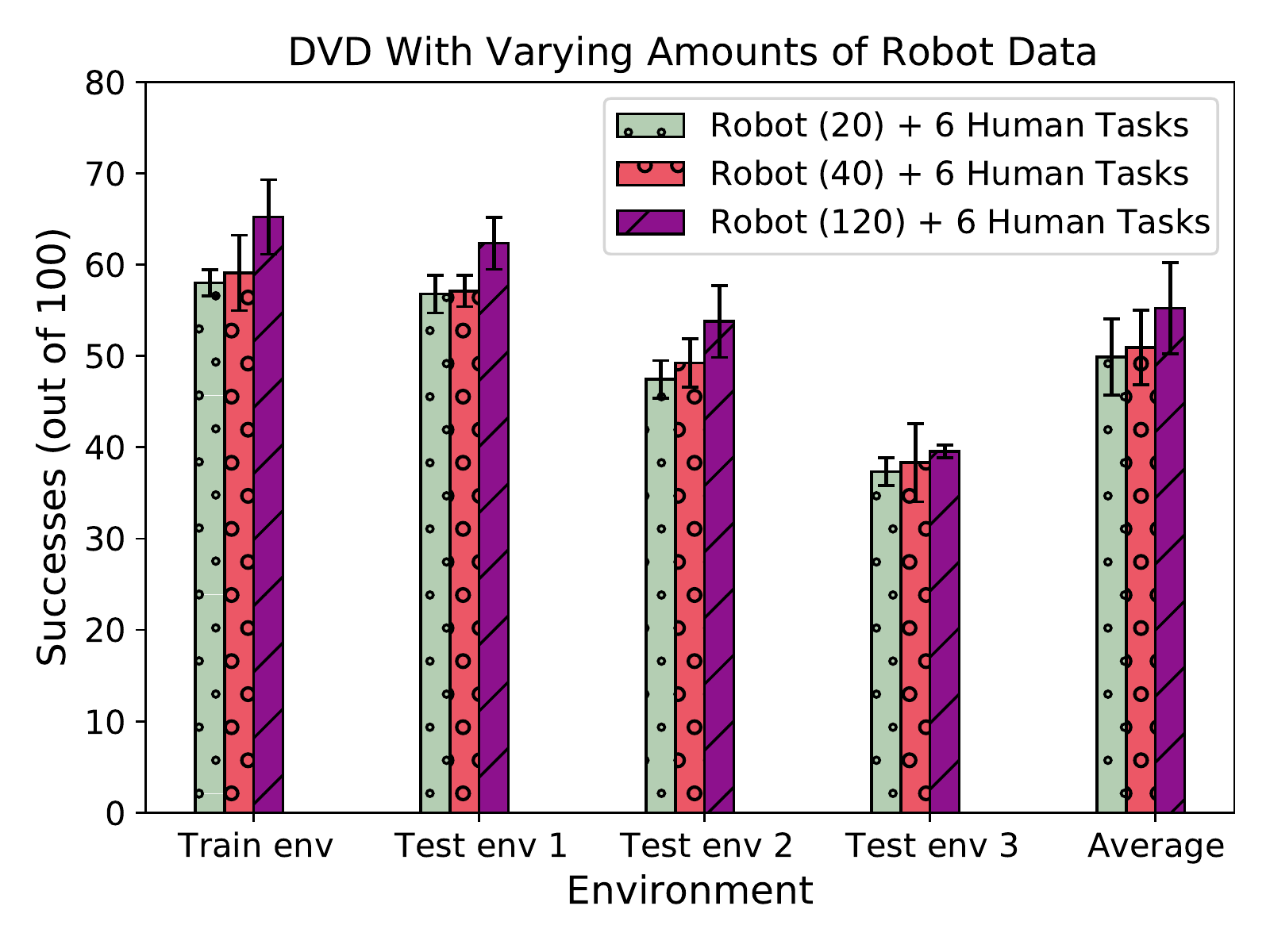}
    \vspace{-0.6cm}
    \caption{\small \textbf{Ablation on Amount of Robot Data Used for Training.} While using 120 robot demonstrations per task slightly benefits performance over using only 20 or 40, \algo still performs comparably with fewer robot demos.}
    \vspace{-0.6cm}
    \label{ablation}
\end{figure}
In Figure~\ref{ablation}, we see that the performance of \algo decreases by only a small margin when using \textbf{as few as 20 robot demonstrations} per task. This suggests that by leveraging the diversity in the human data, \algo can perform well even with very little robot data.

\section{Limitations and Future Work} 
\label{sec:conclusion}

We presented an approach, domain-agnostic video discriminator (\algo),  that leverages the diversity of ``in-the-wild'' human videos to learn generalizable robotic reward functions. Our experiments find that training with a large, diverse dataset of human videos can significantly improve the reward function's ability to generalize to unseen tasks and environments, and can be combined with visual MPC to solve tasks.

There are multiple limitations and directions for future work. First, our method focuses only on learning reward functions that generalize and does not learn a generalizable policy or visual dynamics model directly. This is a necessary next step to achieve agents that broadly generalize and is an exciting direction for future work. Second, while limited in quantity, our work assumes access to some robot demonstrations and task labels for these demos and for all of the human videos. Techniques that can sidestep the need for this supervision would further enhance the scalability of DVD. Lastly, so far we have only tested \algo on coarse tasks that don't require fine-grained manipulation. Designing more powerful visual models and testing \algo~with them on harder, more precise tasks is another exciting direction for future work.

\section*{Acknowledgments}
The authors would like to thank Ashvin Nair as well as members of the IRIS lab for valuable discussions. This work was supported in part by Schmidt Futures, by ONR grant N00014-20-1-2675, and by an NSF GRFP. Chelsea Finn is a CIFAR Fellow in the Learning in Machines \& Brains program.

\bibliographystyle{plainnat}
\bibliography{references}

\begin{thebibliography}{58}
\providecommand{\natexlab}[1]{#1}
\providecommand{\url}[1]{\texttt{#1}}
\expandafter\ifx\csname urlstyle\endcsname\relax
  \providecommand{\doi}[1]{doi: #1}\else
  \providecommand{\doi}{doi: \begingroup \urlstyle{rm}\Url}\fi

\bibitem[Abbeel and Ng(2004)]{apprenticeship_abbeel}
Pieter Abbeel and Andrew~Y. Ng.
\newblock In \emph{Proceedings of the Twenty-First International Conference on
  Machine Learning}, ICML '04, page~1, 2004.

\bibitem[Babaeizadeh et~al.(2018)Babaeizadeh, Finn, Erhan, Campbell, and
  Levine]{babaeizadeh2018stochastic}
Mohammad Babaeizadeh, Chelsea Finn, Dumitru Erhan, Roy~H. Campbell, and Sergey
  Levine.
\newblock Stochastic variational video prediction.
\newblock In \emph{International Conference on Learning Representations}, 2018.

\bibitem[Bonardi et~al.(2020)Bonardi, James, and Davison]{bonardi2019learning}
Alessandro Bonardi, Stephen James, and Andrew~J Davison.
\newblock Learning one-shot imitation from humans without humans.
\newblock \emph{IEEE Robotics and Automation Letters}, 2020.

\bibitem[Brown et~al.(2020)Brown, Mann, Ryder, Subbiah, Kaplan, Dhariwal,
  Neelakantan, Shyam, Sastry, Askell, Agarwal, Herbert-Voss, Krueger, Henighan,
  Child, Ramesh, Ziegler, Wu, Winter, Hesse, Chen, Sigler, Litwin, Gray, Chess,
  Clark, Berner, McCandlish, Radford, Sutskever, and Amodei]{brown2020language}
Tom~B. Brown, Benjamin Mann, Nick Ryder, Melanie Subbiah, Jared Kaplan,
  Prafulla Dhariwal, Arvind Neelakantan, Pranav Shyam, Girish Sastry, Amanda
  Askell, Sandhini Agarwal, Ariel Herbert-Voss, Gretchen Krueger, Tom Henighan,
  Rewon Child, Aditya Ramesh, Daniel~M. Ziegler, Jeffrey Wu, Clemens Winter,
  Christopher Hesse, Mark Chen, Eric Sigler, Mateusz Litwin, Scott Gray,
  Benjamin Chess, Jack Clark, Christopher Berner, Sam McCandlish, Alec Radford,
  Ilya Sutskever, and Dario Amodei.
\newblock Language models are few-shot learners.
\newblock \emph{arXiv:2005.14165}, 2020.

\bibitem[Cabi et~al.(2019)Cabi, G{\'o}mez~Colmenarejo, Novikov, Konyushkova,
  Reed, Jeong, Zolna, Aytar, Budden, Vecerik, et~al.]{cabi2019scaling}
Serkan Cabi, Sergio G{\'o}mez~Colmenarejo, Alexander Novikov, Ksenia
  Konyushkova, Scott Reed, Rae Jeong, Konrad Zolna, Yusuf Aytar, David Budden,
  Mel Vecerik, et~al.
\newblock Scaling data-driven robotics with reward sketching and batch
  reinforcement learning.
\newblock \emph{arXiv:1909.12200}, 2019.

\bibitem[Chang et~al.(2020)Chang, Gupta, and Gupta]{chang2020semantic}
Matthew Chang, Arjun Gupta, and Saurabh Gupta.
\newblock Semantic visual navigation by watching youtube videos.
\newblock In \emph{NeurIPS}, 2020.

\bibitem[Chen et~al.(2021)Chen, Nam, Nair, and Finn]{chen2020batch}
Annie~S. Chen, HyunJi Nam, Suraj Nair, and Chelsea Finn.
\newblock Batch exploration with examples for scalable robotic reinforcement
  learning.
\newblock \emph{IEEE Robotics and Automation Letters}, 2021.

\bibitem[Das et~al.(2021)Das, Bechtle, Davchev, Jayaraman, Rai, and
  Meier]{das2021modelbased}
Neha Das, Sarah Bechtle, Todor Davchev, Dinesh Jayaraman, Akshara Rai, and
  Franziska Meier.
\newblock Model-based inverse reinforcement learning from visual
  demonstrations, 2021.

\bibitem[Das et~al.(2013)Das, Xu, Doell, and Corso]{youcook}
Pradipto Das, Chenliang Xu, Richard~F Doell, and Jason~J Corso.
\newblock A thousand frames in just a few words: Lingual description of videos
  through latent topics and sparse object stitching.
\newblock In \emph{Proceedings of the IEEE conference on computer vision and
  pattern recognition}, pages 2634--2641, 2013.

\bibitem[Dasari et~al.(2019)Dasari, Ebert, Tian, Nair, Bucher, Schmeckpeper,
  Singh, Levine, and Finn]{dasari2019robonet}
Sudeep Dasari, Frederik Ebert, Stephen Tian, Suraj Nair, Bernadette Bucher,
  Karl Schmeckpeper, Siddharth Singh, Sergey Levine, and Chelsea Finn.
\newblock Robonet: Large-scale multi-robot learning.
\newblock In \emph{Conference on Robot Learning}, 2019.

\bibitem[Deng et~al.(2009)Deng, Dong, Socher, Li, Li, and
  Fei-Fei]{imagenet_cvpr09}
J.~Deng, W.~Dong, R.~Socher, L.-J. Li, K.~Li, and L.~Fei-Fei.
\newblock {ImageNet: A Large-Scale Hierarchical Image Database}.
\newblock In \emph{CVPR09}, 2009.

\bibitem[Devlin et~al.(2019)Devlin, Chang, Lee, and Toutanova]{bert}
Jacob Devlin, Ming-Wei Chang, Kenton Lee, and Kristina Toutanova.
\newblock {BERT}: Pre-training of deep bidirectional transformers for language
  understanding.
\newblock In \emph{Conference of the North American Chapter of the Association
  for Computational Linguistics: Human Language Technologies (NAACL-HLT)},
  Minneapolis, Minnesota, June 2019. Association for Computational Linguistics.

\bibitem[Ebert et~al.(2018)Ebert, Finn, Dasari, Xie, Lee, and
  Levine]{ebert2018visual}
Frederik Ebert, Chelsea Finn, Sudeep Dasari, Annie Xie, Alex Lee, and Sergey
  Levine.
\newblock Visual foresight: Model-based deep reinforcement learning for
  vision-based robotic control.
\newblock \emph{arXiv:1812.00568}, 2018.

\bibitem[Edwards and Isbell(2019)]{edwards2019perceptual}
Ashley~D Edwards and Charles~L Isbell.
\newblock Perceptual values from observation.
\newblock \emph{arXiv preprint arXiv:1905.07861}, 2019.

\bibitem[Fabian Caba~Heilbron and Niebles(2015)]{caba2015activitynet}
Bernard~Ghanem Fabian Caba~Heilbron, Victor~Escorcia and Juan~Carlos Niebles.
\newblock Activitynet: A large-scale video benchmark for human activity
  understanding.
\newblock In \emph{Proceedings of the IEEE Conference on Computer Vision and
  Pattern Recognition}, pages 961--970, 2015.

\bibitem[Finn and Levine(2017)]{finn2017deep}
Chelsea Finn and Sergey Levine.
\newblock Deep visual foresight for planning robot motion.
\newblock In \emph{IEEE International Conference on Robotics and Automation
  (ICRA)}, 2017.

\bibitem[Finn et~al.(2016)Finn, Levine, and Abbeel]{finn2016guided}
Chelsea Finn, Sergey Levine, and Pieter Abbeel.
\newblock Guided cost learning: Deep inverse optimal control via policy
  optimization.
\newblock In \emph{International conference on machine learning}, pages 49--58.
  PMLR, 2016.

\bibitem[Fu et~al.(2018{\natexlab{a}})Fu, Luo, and Levine]{fu2018learning}
Justin Fu, Katie Luo, and Sergey Levine.
\newblock Learning robust rewards with adverserial inverse reinforcement
  learning.
\newblock In \emph{International Conference on Learning Representations},
  2018{\natexlab{a}}.

\bibitem[Fu et~al.(2018{\natexlab{b}})Fu, Singh, Ghosh, Yang, and
  Levine]{fu2018variational}
Justin Fu, Avi Singh, Dibya Ghosh, Larry Yang, and Sergey Levine.
\newblock Variational inverse control with events: A general framework for
  data-driven reward definition.
\newblock In \emph{Advances in Neural Information Processing Systems},
  2018{\natexlab{b}}.

\bibitem[{Goo} and {Niekum}(2019)]{goo_2019_multistep}
W.~{Goo} and S.~{Niekum}.
\newblock One-shot learning of multi-step tasks from observation via activity
  localization in auxiliary video.
\newblock In \emph{2019 International Conference on Robotics and Automation
  (ICRA)}, 2019.

\bibitem[Goyal et~al.(2017)Goyal, Ebrahimi~Kahou, Michalski, Materzynska,
  Westphal, Kim, Haenel, Fruend, Yianilos, Mueller-Freitag,
  et~al.]{goyal2017something}
Raghav Goyal, Samira Ebrahimi~Kahou, Vincent Michalski, Joanna Materzynska,
  Susanne Westphal, Heuna Kim, Valentin Haenel, Ingo Fruend, Peter Yianilos,
  Moritz Mueller-Freitag, et~al.
\newblock The" something something" video database for learning and evaluating
  visual common sense.
\newblock In \emph{Proceedings of the IEEE International Conference on Computer
  Vision}, pages 5842--5850, 2017.

\bibitem[Gupta et~al.(2018)Gupta, Murali, Gandhi, and Pinto]{gupta2018robot}
Abhinav Gupta, Adithyavairavan Murali, Dhiraj~Prakashchand Gandhi, and Lerrel
  Pinto.
\newblock Robot learning in homes: Improving generalization and reducing
  dataset bias.
\newblock In \emph{Advances in Neural Information Processing Systems}, 2018.

\bibitem[Hafner et~al.(2019)Hafner, Lillicrap, Fischer, Villegas, Ha, Lee, and
  Davidson]{hafner2019learning}
Danijar Hafner, Timothy Lillicrap, Ian Fischer, Ruben Villegas, David Ha,
  Honglak Lee, and James Davidson.
\newblock Learning latent dynamics for planning from pixels.
\newblock In \emph{International Conference on Machine Learning}, pages
  2555--2565. PMLR, 2019.

\bibitem[Kalashnikov et~al.(2018)Kalashnikov, Irpan, Pastor, Ibarz, Herzog,
  Jang, Quillen, Holly, Kalakrishnan, Vanhoucke, et~al.]{kalashnikov2018qtopt}
Dmitry Kalashnikov, Alex Irpan, Peter Pastor, Julian Ibarz, Alexander Herzog,
  Eric Jang, Deirdre Quillen, Ethan Holly, Mrinal Kalakrishnan, Vincent
  Vanhoucke, et~al.
\newblock Scalable deep reinforcement learning for vision-based robotic
  manipulation.
\newblock In \emph{Conference on Robot Learning}, pages 651--673. PMLR, 2018.

\bibitem[Lee and Ryoo(2017)]{lee2017learning}
Jangwon Lee and Michael~S Ryoo.
\newblock Learning robot activities from first-person human videos using
  convolutional future regression.
\newblock In \emph{Proceedings of the IEEE Conference on Computer Vision and
  Pattern Recognition Workshops}, pages 1--2, 2017.

\bibitem[Lee et~al.(2013)Lee, Su, Kim, and Demiris]{lee_2013_syntactic}
Kyuhwa Lee, Yanyu Su, Tae-Kyun Kim, and Yiannis Demiris.
\newblock A syntactic approach to robot imitation learning using probabilistic
  activity grammars.
\newblock \emph{Robotics and Autonomous Systems}, 61\penalty0 (12):\penalty0
  1323--1334, 2013.

\bibitem[Liu et~al.(2018)Liu, Gupta, Abbeel, and Levine]{liu2018imitation}
YuXuan Liu, Abhishek Gupta, Pieter Abbeel, and Sergey Levine.
\newblock Imitation from observation: Learning to imitate behaviors from raw
  video via context translation.
\newblock In \emph{2018 IEEE International Conference on Robotics and
  Automation (ICRA)}, pages 1118--1125. IEEE, 2018.

\bibitem[Mandlekar et~al.(2018)Mandlekar, Zhu, Garg, Booher, Spero, Tung, Gao,
  Emmons, Gupta, Orbay, Savarese, and Fei-Fei]{mandlekar2018roboturk}
Ajay Mandlekar, Yuke Zhu, Animesh Garg, Jonathan Booher, Max Spero, Albert
  Tung, Julian Gao, John Emmons, Anchit Gupta, Emre Orbay, Silvio Savarese, and
  Li~Fei-Fei.
\newblock Roboturk: A crowdsourcing platform for robotic skill learning through
  imitation.
\newblock In \emph{Conference on Robot Learning}, 2018.

\bibitem[Nair et~al.(2018)Nair, Pong, Dalal, Bahl, Lin, and
  Levine]{nair2018visual}
Ashvin~V Nair, Vitchyr Pong, Murtaza Dalal, Shikhar Bahl, Steven Lin, and
  Sergey Levine.
\newblock Visual reinforcement learning with imagined goals.
\newblock In \emph{Advances in Neural Information Processing Systems}, 2018.

\bibitem[Nguyen et~al.(2018)Nguyen, Kanoulas, Muratore, Caldwell, and
  Tsagarakis]{nguyen2017translating}
Anh Nguyen, Dimitrios Kanoulas, Luca Muratore, Darwin~G Caldwell, and Nikos~G
  Tsagarakis.
\newblock Translating videos to commands for robotic manipulation with deep
  recurrent neural networks.
\newblock In \emph{2018 IEEE International Conference on Robotics and
  Automation (ICRA)}, pages 3782--3788. IEEE, 2018.

\bibitem[OpenAI et~al.(2019)OpenAI, Andrychowicz, Baker, Chociej, Jozefowicz,
  McGrew, Pachocki, Petron, Plappert, Powell, Ray, Schneider, Sidor, Tobin,
  Welinder, Weng, and Zaremba]{openai2019learning}
OpenAI, Marcin Andrychowicz, Bowen Baker, Maciek Chociej, Rafal Jozefowicz, Bob
  McGrew, Jakub Pachocki, Arthur Petron, Matthias Plappert, Glenn Powell, Alex
  Ray, Jonas Schneider, Szymon Sidor, Josh Tobin, Peter Welinder, Lilian Weng,
  and Wojciech Zaremba.
\newblock Learning dexterous in-hand manipulation, 2019.

\bibitem[Petrík et~al.(2020)Petrík, Tapaswi, Laptev, and
  Sivic]{petrik2020learning}
Vladimír Petrík, Makarand Tapaswi, Ivan Laptev, and Josef Sivic.
\newblock Learning object manipulation skills via approximate state estimation
  from real videos, 2020.

\bibitem[Pinto and Gupta(2016)]{pinto2016supersizing}
Lerrel Pinto and Abhinav Gupta.
\newblock Supersizing self-supervision: Learning to grasp from 50k tries and
  700 robot hours.
\newblock In \emph{IEEE international conference on robotics and automation
  (ICRA)}, 2016.

\bibitem[Pirk et~al.(2019)Pirk, Khansari, Bai, Lynch, and
  Sermanet]{pirk2019online}
Sören Pirk, Mohi Khansari, Yunfei Bai, Corey Lynch, and Pierre Sermanet.
\newblock Online object representations with contrastive learning, 2019.

\bibitem[Ratliff et~al.(2006)Ratliff, Bagnell, and
  Zinkevich]{ratcliff_maxmargin}
Nathan~D. Ratliff, J.~Andrew Bagnell, and Martin~A. Zinkevich.
\newblock Maximum margin planning.
\newblock In \emph{Proceedings of the 23rd International Conference on Machine
  Learning}, ICML '06, page 729–736, 2006.

\bibitem[Rothfuss et~al.(2018)Rothfuss, Ferreira, Aksoy, Zhou, and
  Asfour]{rothfuss2018deep}
Jonas Rothfuss, Fabio Ferreira, Eren~Erdal Aksoy, You Zhou, and Tamim Asfour.
\newblock Deep episodic memory: Encoding, recalling, and predicting episodic
  experiences for robot action execution.
\newblock \emph{IEEE Robotics and Automation Letters}, 3\penalty0 (4):\penalty0
  4007--4014, 2018.

\bibitem[Rubinstein and Kroese(2013)]{rubinstein2013cross}
Reuven~Y Rubinstein and Dirk~P Kroese.
\newblock \emph{The cross-entropy method: a unified approach to combinatorial
  optimization, Monte-Carlo simulation and machine learning}.
\newblock Springer Science \& Business Media, 2013.

\bibitem[Scalise et~al.(2019)Scalise, Thomason, Bisk, and
  Srinivasa]{scalise2019improving}
Rosario Scalise, Jesse Thomason, Yonatan Bisk, and Siddhartha Srinivasa.
\newblock Improving robot success detection using static object data.
\newblock In \emph{Proceedings of the 2019 IEEE/RSJ International Conference on
  Intelligent Robots and Systems}, 2019.

\bibitem[Schmeckpeper et~al.(2020{\natexlab{a}})Schmeckpeper, Rybkin,
  Daniilidis, Levine, and Finn]{schmeckpeper2020reinforcement}
Karl Schmeckpeper, Oleh Rybkin, Kostas Daniilidis, Sergey Levine, and Chelsea
  Finn.
\newblock Reinforcement learning with videos: Combining offline observations
  with interaction.
\newblock In \emph{CoRL}, 2020{\natexlab{a}}.

\bibitem[Schmeckpeper et~al.(2020{\natexlab{b}})Schmeckpeper, Xie, Rybkin,
  Tian, Daniilidis, Levine, and Finn]{schmeckpeper2019learning}
Karl Schmeckpeper, Annie Xie, Oleh Rybkin, Stephen Tian, Kostas Daniilidis,
  Sergey Levine, and Chelsea Finn.
\newblock Learning predictive models from observation and interaction.
\newblock In \emph{ECCV}, 2020{\natexlab{b}}.

\bibitem[Sermanet et~al.(2017)Sermanet, Xu, and
  Levine]{sermanet2017unsupervised}
Pierre Sermanet, Kelvin Xu, and Sergey Levine.
\newblock Unsupervised perceptual rewards for imitation learning.
\newblock \emph{Proceedings of Robotics: Science and Systems (RSS)}, 2017.

\bibitem[Sermanet et~al.(2018)Sermanet, Lynch, Chebotar, Hsu, Jang, Schaal, and
  Levine]{sermanet2018timecontrastive}
Pierre Sermanet, Corey Lynch, Yevgen Chebotar, Jasmine Hsu, Eric Jang, Stefan
  Schaal, and Sergey Levine.
\newblock Time-contrastive networks: Self-supervised learning from video.
\newblock \emph{Proceedings of International Conference in Robotics and
  Automation (ICRA)}, 2018.

\bibitem[Shao et~al.(2020)Shao, Migimatsu, Zhang, Yang, and
  Bohg]{shao2020concept}
Lin Shao, Toki Migimatsu, Qiang Zhang, Karen Yang, and Jeannette Bohg.
\newblock Concept2robot: Learning manipulation concepts from instructions and
  human demonstrations.
\newblock In \emph{Proceedings of Robotics: Science and Systems (RSS)}, 2020.

\bibitem[Sharma et~al.(2019)Sharma, Pathak, and Gupta]{sharma2019thirdperson}
P.~Sharma, Deepak Pathak, and Abhinav Gupta.
\newblock Third-person visual imitation learning via decoupled hierarchical
  controller.
\newblock In \emph{NeurIPS}, 2019.

\bibitem[Singh et~al.(2019)Singh, Yang, Finn, and Levine]{singh2019end}
Avi Singh, Larry Yang, Chelsea Finn, and Sergey Levine.
\newblock End-to-end robotic reinforcement learning without reward engineering.
\newblock In \emph{Proceedings of Robotics: Science and Systems},
  FreiburgimBreisgau, Germany, June 2019.

\bibitem[Singh et~al.(2020)Singh, Jang, Irpan, Kappler, Dalal, Levinev,
  Khansari, and Finn]{singh2020scalable}
Avi Singh, Eric Jang, Alexander Irpan, Daniel Kappler, Murtaza Dalal, Sergey
  Levinev, Mohi Khansari, and Chelsea Finn.
\newblock Scalable multi-task imitation learning with autonomous improvement.
\newblock In \emph{2020 IEEE International Conference on Robotics and
  Automation (ICRA)}, pages 2167--2173. IEEE, 2020.

\bibitem[Smith et~al.(2020)Smith, Dhawan, Zhang, Abbeel, and
  Levine]{smith2020avid}
Laura Smith, Nikita Dhawan, Marvin Zhang, Pieter Abbeel, and Sergey Levine.
\newblock {AVID: Learning Multi-Stage Tasks via Pixel-Level Translation of
  Human Videos}.
\newblock In \emph{Proceedings of Robotics: Science and Systems}, Corvalis,
  Oregon, USA, July 2020.

\bibitem[{Todorov} et~al.(2012){Todorov}, {Erez}, and {Tassa}]{mujoco}
E.~{Todorov}, T.~{Erez}, and Y.~{Tassa}.
\newblock Mujoco: A physics engine for model-based control.
\newblock In \emph{2012 IEEE/RSJ International Conference on Intelligent Robots
  and Systems}, 2012.

\bibitem[Watter et~al.(2015)Watter, Springenberg, Boedecker, and
  Riedmiller]{watter2015embed}
Manuel Watter, Jost~Tobias Springenberg, Joschka Boedecker, and Martin
  Riedmiller.
\newblock Embed to control: a locally linear latent dynamics model for control
  from raw images.
\newblock In \emph{Proceedings of the 28th International Conference on Neural
  Information Processing Systems-Volume 2}, pages 2746--2754, 2015.

\bibitem[Wulfmeier et~al.(2016)Wulfmeier, Ondruska, and
  Posner]{wulfmeier2016maximum}
Markus Wulfmeier, Peter Ondruska, and Ingmar Posner.
\newblock Maximum entropy deep inverse reinforcement learning, 2016.

\bibitem[Xiong et~al.(2021)Xiong, Li, Chen, Bharadhwaj, Sinha, and
  Garg]{xiong2021learning}
Haoyu Xiong, Quanzhou Li, Yun-Chun Chen, Homanga Bharadhwaj, Samarth Sinha, and
  Animesh Garg.
\newblock Learning by watching: Physical imitation of manipulation skills from
  human videos, 2021.

\bibitem[Yang et~al.(2015)Yang, Li, Fermüller, and Aloimonos]{yang_robot_WWW}
Yezhou Yang, Yi~Li, Cornelia Fermüller, and Yiannis Aloimonos.
\newblock Robot learning manipulation action plans by "watching" unconstrained
  videos from the world wide web.
\newblock In \emph{AAAI}, pages 3686--3693, 2015.

\bibitem[Young et~al.(2020)Young, Gandhi, Tulsiani, Gupta, Abbeel, and
  Pinto]{young2020visual}
Sarah Young, Dhiraj Gandhi, Shubham Tulsiani, Abhinav Gupta, Pieter Abbeel, and
  Lerrel Pinto.
\newblock Visual imitation made easy.
\newblock In \emph{CoRL}, 2020.

\bibitem[Yu et~al.(2018)Yu, Finn, Dasari, Xie, Zhang, Abbeel, and Levine]{daml}
Tianhe Yu, Chelsea Finn, Sudeep Dasari, Annie Xie, Tianhao Zhang, Pieter
  Abbeel, and Sergey Levine.
\newblock One-shot imitation from observing humans via domain-adaptive
  meta-learning.
\newblock In \emph{Proceedings of Robotics: Science and Systems}, Pittsburgh,
  Pennsylvania, June 2018.

\bibitem[Yu et~al.(2020)Yu, Quillen, He, Julian, Hausman, Finn, and
  Levine]{yu2020meta}
Tianhe Yu, Deirdre Quillen, Zhanpeng He, Ryan Julian, Karol Hausman, Chelsea
  Finn, and Sergey Levine.
\newblock Meta-world: A benchmark and evaluation for multi-task and meta
  reinforcement learning.
\newblock In \emph{Conference on Robot Learning}, 2020.

\bibitem[Zeng et~al.(2018)Zeng, Song, Welker, Lee, Rodriguez, and
  Funkhouser]{zeng2018learning}
Andy Zeng, Shuran Song, Stefan Welker, Johnny Lee, Alberto Rodriguez, and
  Thomas Funkhouser.
\newblock Learning synergies between pushing and grasping with self-supervised
  deep reinforcement learning.
\newblock \emph{Proceedings of the IEEE International Conference on Intelligent
  Robots and Systems (IROS)}, 2018.

\bibitem[Zhu et~al.(2020)Zhu, Yu, Gupta, Shah, Hartikainen, Singh, Kumar, and
  Levine]{Zhu2020The}
Henry Zhu, Justin Yu, Abhishek Gupta, Dhruv Shah, Kristian Hartikainen, Avi
  Singh, Vikash Kumar, and Sergey Levine.
\newblock The ingredients of real world robotic reinforcement learning.
\newblock In \emph{International Conference on Learning Representations}, 2020.

\bibitem[Ziebart et~al.(2008)Ziebart, Maas, Bagnell, and
  Dey]{ziebart2008maximum}
Brian~D. Ziebart, Andrew Maas, J.~Andrew Bagnell, and Anind~K. Dey.
\newblock Maximum entropy inverse reinforcement learning.
\newblock In \emph{Proc. AAAI}, pages 1433--1438, 2008.

\end{thebibliography}

\clearpage
\appendix

\subsection{Training Details}
\label{sec:appendix-training}

\paragraph{Dataset Details}
Depending on the experiment, we choose videos from the following 15 different human tasks in the Something-Something-V2 dataset for training \algo, where each task has from 853-3170 training videos:
1) Closing sth, 2) Moving sth away from camera, 3) Moving sth towards camera, 4) Opening sth, 5) Pushing sth left to right, 6) Pushing sth right to left, 7) Poking sth so lightly it doesn't move, 8) Moving sth down, 9) Moving sth up, 10) Pulling sth from left to right, 11) Pulling sth from right to left, 12) Pushing sth with sth, 13) Moving sth closer to sth, 14) Plugging sth into sth, and 15) Pushing sth so that it slightly moves. We used these tasks because they are appropriate for a single-arm setting and cover a diverse range of various actions. The first seven of those tasks were chosen as they are relevant to tasks possible in the simulation environments, but the other tasks were not chosen for any particular reason, i.e. could have been replaced by a different set of 8 other appropriate tasks. For an experiment, if human videos for a task are used, we use \emph{all} of the human videos available in the \sth training set for that task to train \algo. 

For the simulation experiments, \algo is also trained on 120 robot video demonstrations of 3 tasks, half of which are collected in the original training environment and the other half in the rearranged training environment. These videos are collected via model predictive control with random shooting, a ground truth video-prediction model, and a ground-truth shaped reward particular to each task. 
For the real robot experiments on the WidowX200, in addition to varying amounts of human videos, \algo is trained on 80 robot video demonstrations of 2 tasks, which are collected in the original training environment. These demonstrations are collected via a hard-coded script with uniform noise between -0.02 and 0.02 added to each action. 

To evaluate \algo's training progress, we use a validation set consisting of all of the human videos available in the \sth validation set for the chosen tasks as well as 48 robot video demonstrations for the same 3 tasks with robot demos in the training set, with half of these coming from the original training environment and the other half from the rearranged. For the WidowX200 experiments, we add 8 robot video demonstrations for each of the 2 tasks into the validation set. 

\paragraph{Hyperparameters}

For \algo, the similarity discriminator is trained with a learning rate of 0.01 using stochastic gradient descent (SGD) with momentum 0.9 and weight decay 0.00001. We use a batch size of 24, where each element of the batch consists of a triplet with two videos having the same task label and the third having a different label. Each version of \algo in the experiments is trained for 120 epochs, where one epoch consists of 200 optimizer steps. For each epoch, the video clips fed into \algo for  training are sequences of consecutive frames with random length between 20 and 40 frames taken from the original video. If the original video has fewer than the randomly selected amount of frames, the last frame is repeated to achieve the desired number of frames for the training clip. Additionally, during training, each input video is first randomly rotated between -15 and 15 degrees, scaled so that the height has size 120, and then randomly cropped to have size $120 \times 120 \times 3$. At planning time, the video demonstration is spliced so that it is between 30 and 40 frames, rescaled to have a height of 120 pixels, and then center-cropped to have size $120 \times 120 \times 3$.
The demo-conditioned behavioral cloning baseline uses the same hyperparameters and training details except that it uses weight decay 0.0001.

\begin{figure}%
    \centering
    \vspace{0.1cm}
    \includegraphics[width=0.99\linewidth]{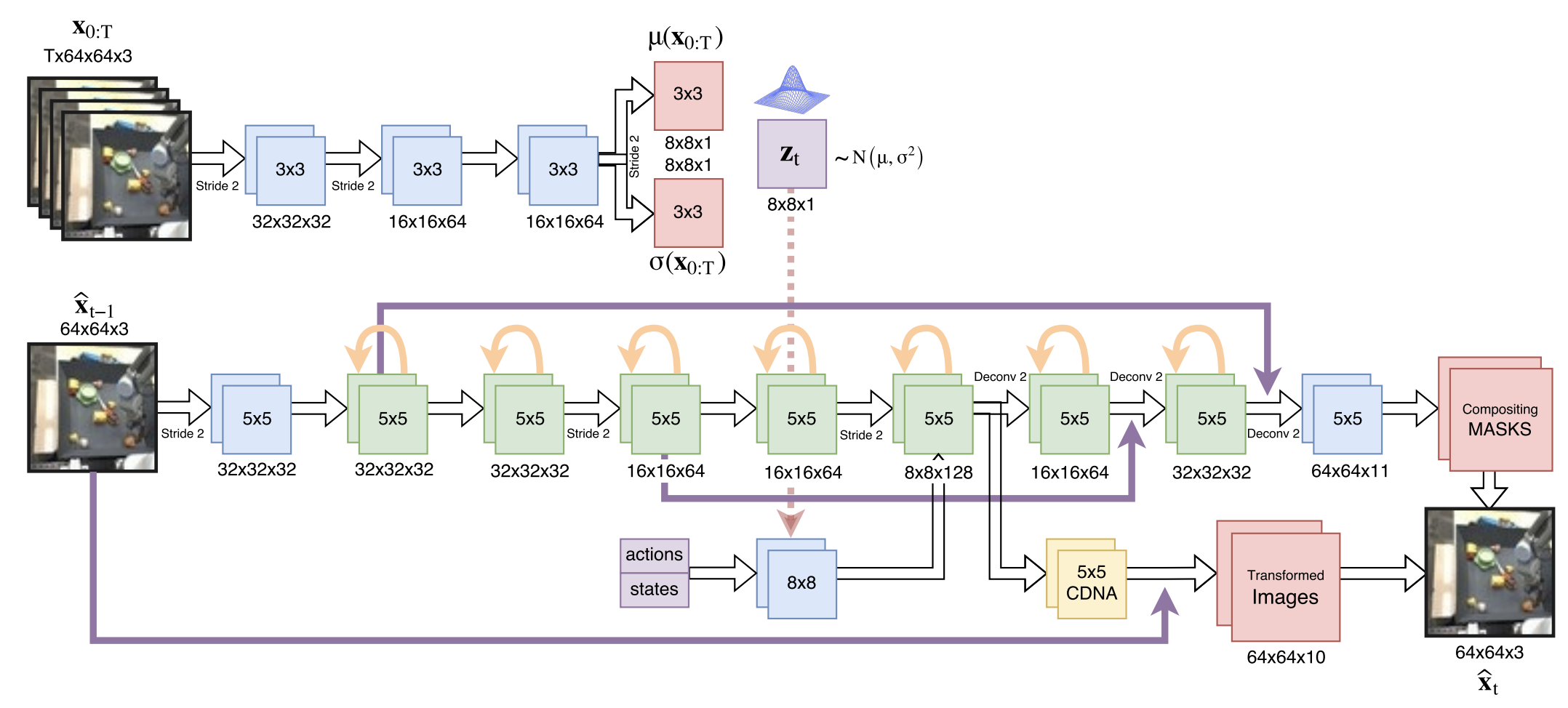}
    \vspace{-0.3cm}
    \caption{\small \textbf{SV2P Architecture.} We use video prediction models trained via SV2P with the reward from \algo in order to complete tasks specified by a given human video. Figure taken from the original paper~\cite{babaeizadeh2018stochastic}.}
    \vspace{-0.6cm}
    \label{fig:sv2p}
\end{figure}

\paragraph{SV2P Training}
To evaluate \algo's performance on potentially unseen tasks with a given human video, we employ visual model predictive control with SV2P visual prediction models trained on datasets autonomously collected in each environment. SV2P learns an action-conditioned video prediction model by sampling a latent variable and subsequently generating an image prediction with that sample. We use the same architecture, which is shown in Figure~\ref{fig:sv2p}, and default hyperparameters as the original paper~\cite{babaeizadeh2018stochastic}. 

For each of the four simulation environments in which we evaluate \algo, we collect 10,000 random episodes, each with 60 total frames, of the agent interacting in that environment and train SV2P for 200,000 epochs on all of the data. The models are trained to predict the next fifteen frames given an input of five frames. To evaluate \algo in the robot test environment, we train SV2P for 160,000 epochs on 58,500 frames worth of autonomously collected robot interaction in the original train environment, and then we finetune the model for another 60,000 epochs on 21,000 frames worth of autonomously collected data in the test environment that has the toy kitchen door. Collecting this data on the WidowX200 took a total of roughly 75 hours but was entirely autonomous. 

\paragraph{Additional \algo Details}
Here we expand on some of the \algo implementation details touched upon in Section~\ref{sec:method-DVD-implementation}, particularly the way that batches are sampled during training. Each batch consists of triplets $(d_i, d'_i, d_j)$, where $d_j$ is labeled as a different task as $d_i$ and $d'_i$, which are labeled as the same task. 
In each triplet, $d_i$ is randomly sampled with 0.5 probability of being a robot demonstration. Then, if $d_i$ is from a task with only human data, $d'_i$ will be chosen from the remaining human data for that task; otherwise it is chosen to be a robot video from that task with 0.5 probability. Finally, $d_j$ is randomly sampled repeatedly (usually just once) with 0.5 probability of being a robot demonstration until a video with a different task label from $d_i$ and $d'_i$ is sampled.

\paragraph{Comparisons}
For the Concept2Robot comparison, we use the same 174-way classifier that the paper used and do not alter it. For the demo-conditioned behavioral cloning comparison, we use a method similar to~\cite{bonardi2019learning},~\cite{singh2020scalable}, and~\cite{daml}. We train a model that takes in as input the concatenated encodings of a conditioning video and the image state from one of the robot demonstrations in the training set and outputs an action that aims to lead the agent from the given image state to completing the same task as shown in the conditioning demo. During training, the model is trained on batches of (conditioning video, robot demonstration) pairs, where the conditioning video is randomly taken from the combined human and robot dataset and a robot demonstration with the same task label is randomly chosen. Because there are many more human videos than robot demonstrations, the conditioning video is chosen to be a robot demonstration with 50\% probability, which is analogous to the balancing of batches used in \algo. Note that this method cannot naturally use human videos from tasks for which there are no robot demonstrations. 

The behavior cloning model uses the same pretrained video encoder as \algo to encode the conditioning demo as well as a pretrained ResNet18 for the image state. The resulting features are concatenated and passed into an MLP that takes an input of size [1512] and has fully connected layers $[512, 256, 128, 64, 32, a]$, where each layer except the last is followed by a ReLU activation and $a$ corresponds to the number of action dimensions. The model is trained to minimize mean squared error between the output action and the true action.

\subsection{Experimental Details}
\label{sec:appendix-exp-details}

\paragraph{Domains} For the simulation domains, we use a Mujoco simulation built off the Meta-World environments \cite{yu2020meta}. In simulation, the state space is the space of RGB image observations with size [180, 120, 3]. We use a continuous action space over the linear and angular velocity of the robot’s gripper and a discrete action space over the gripper open/close action, for a total of five dimensions. For the robot domain, we consider a real WidowX200 robot interacting with a file cabinet, a tissue box, a stuffed animal, and a toy kitchen set. The state space is the space of RGB image observations with size [120, 120, 3], and the action space consists of the continuous linear velocity of the robot's gripper in the x and z directions as well as the gripper's y-position, for a total of three dimensions. 

\paragraph{Simulation Experiments}
For all environment and task generalization experiments, in each trial we plan 3 trajectories of length 20. For each trajectory, we sample 100 action sequences uniformly randomly and randomly choose one of the top 5 predicted trajectories with the highest functional similarity score given by \algo to execute in the environment. For Concept2Robot, we take one of the top 5 predicted trajectories with the highest classification score for the specified task, and for the behavioral cloning policy, we simply take the predicted action at each state. 
We evaluate on the following three target tasks: 1) Closing the drawer, which is defined as the last frame in the 60-frame trajectory having the drawer pushed in to be less than 0.05, where it starts open at 0.07, 2) Turning the faucet to the right more than 0.01 distance, where it starts at 0, and 3) Moves cup to be less than 0.07 distance to the coffee machine, where the cup starts out at least 0.1 away. We run 100 trials for 3 different seeds for each task for every method in all experiments. 

\paragraph{Real Robot Experiments}
On the WidowX200, for all experiments, in each trial we plan 1 trajectory of length 10. For this trajectory, we run 2 iterations of the cross-entropy method (CEM), sampling 100 action sequences and refitting to the top 20 repeatedly. We then choose one of the top 5 predicted trajectories with the highest functional similarity score given by \algo to execute in the environment. 
We evaluate on the following two target tasks: 1) Closing the toy kitchen door, where a success is recorded for any trial where the robot arm completely closes the door, and 2) Pushing the tissue box to the left, where the robot arm must clearly push the tissue box left of its original starting position. We run 20 trials for each task for each method.

\subsection{Additional Experimental Results}
\label{sec:appendix-exp-results}

\begin{figure}%
    \centering
    \vspace{0.1cm}
    \includegraphics[width=0.99\linewidth]{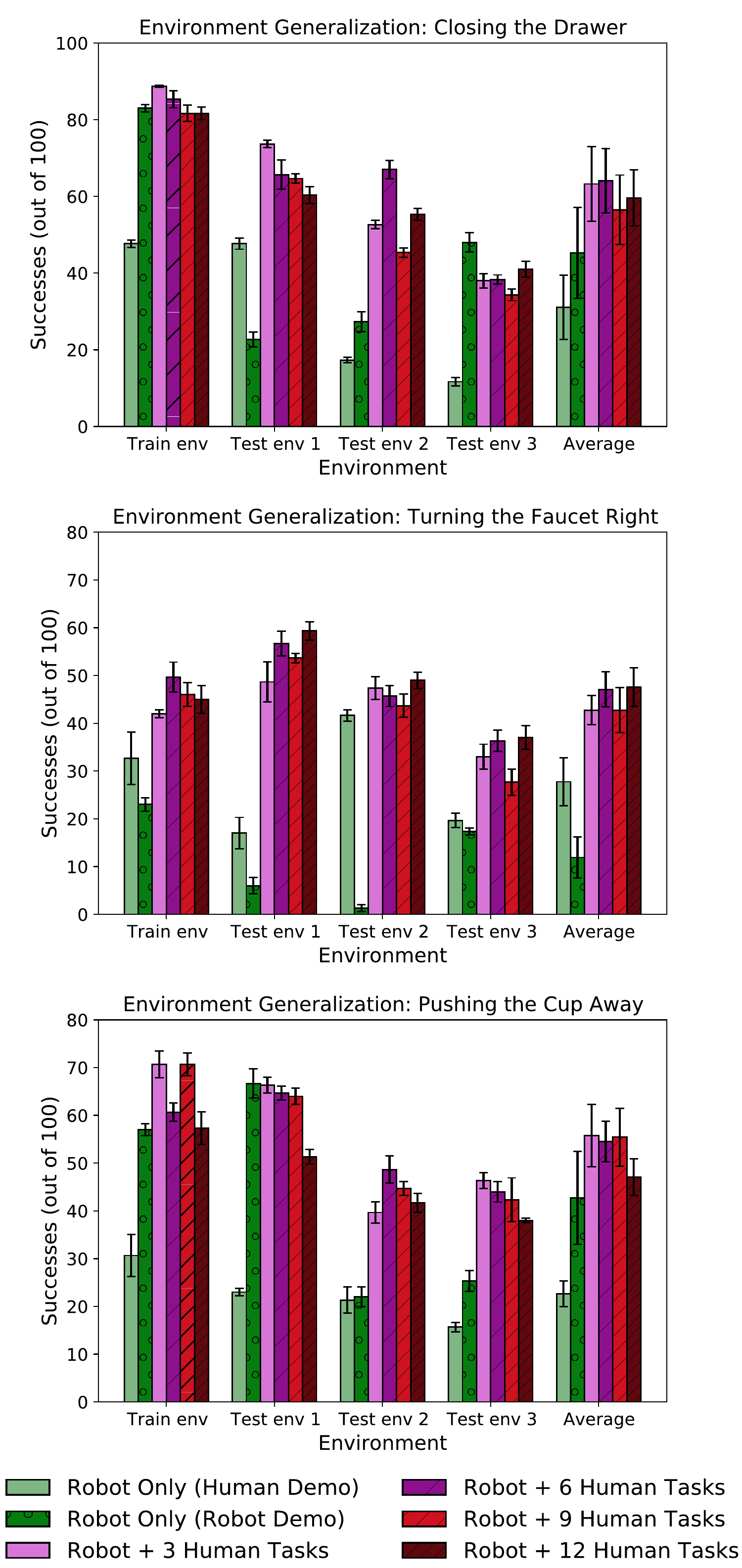}
    \vspace{-0.3cm}
    \caption{\small \textbf{Effect of Human Data on Environment Generalization.} We compare \algo's performance on seen and unseen environments when trained on only robot videos compared to varying number of human videos. Across all three tasks, we see that training with human videos provides significantly improved performance over only training on robot videos, and that \algo is generally robust to the number of different human video tasks used. Each bar shows the average success rate over all 3 target tasks, computed over 3 seeds of 100 trials, with error bars denoting standard error.}
    \vspace{-0.6cm}
    \label{fig:appendix-env-gen}
\end{figure}

In our simulation environment generalization experiments, we evaluate on the three tasks of 1) Closing the drawer, 2) Turning the faucet to the right, and 3) Pushing the cup away from the camera. In Section~\ref{sec:exps}, we reported the average performance across all the three tasks. In Figure~\ref{fig:appendix-env-gen}, we present the individual task results for DVD trained with varying amounts of human data. The conclusions of these experiments are the same as those in Section~\ref{sec:exps}, in that leveraging diverse human videos in \algo allows for more effective generalization across new environments rather than relying only on robot videos. 

\begin{figure}%
    \centering
    \vspace{0.1cm}
    \includegraphics[width=0.9\linewidth]{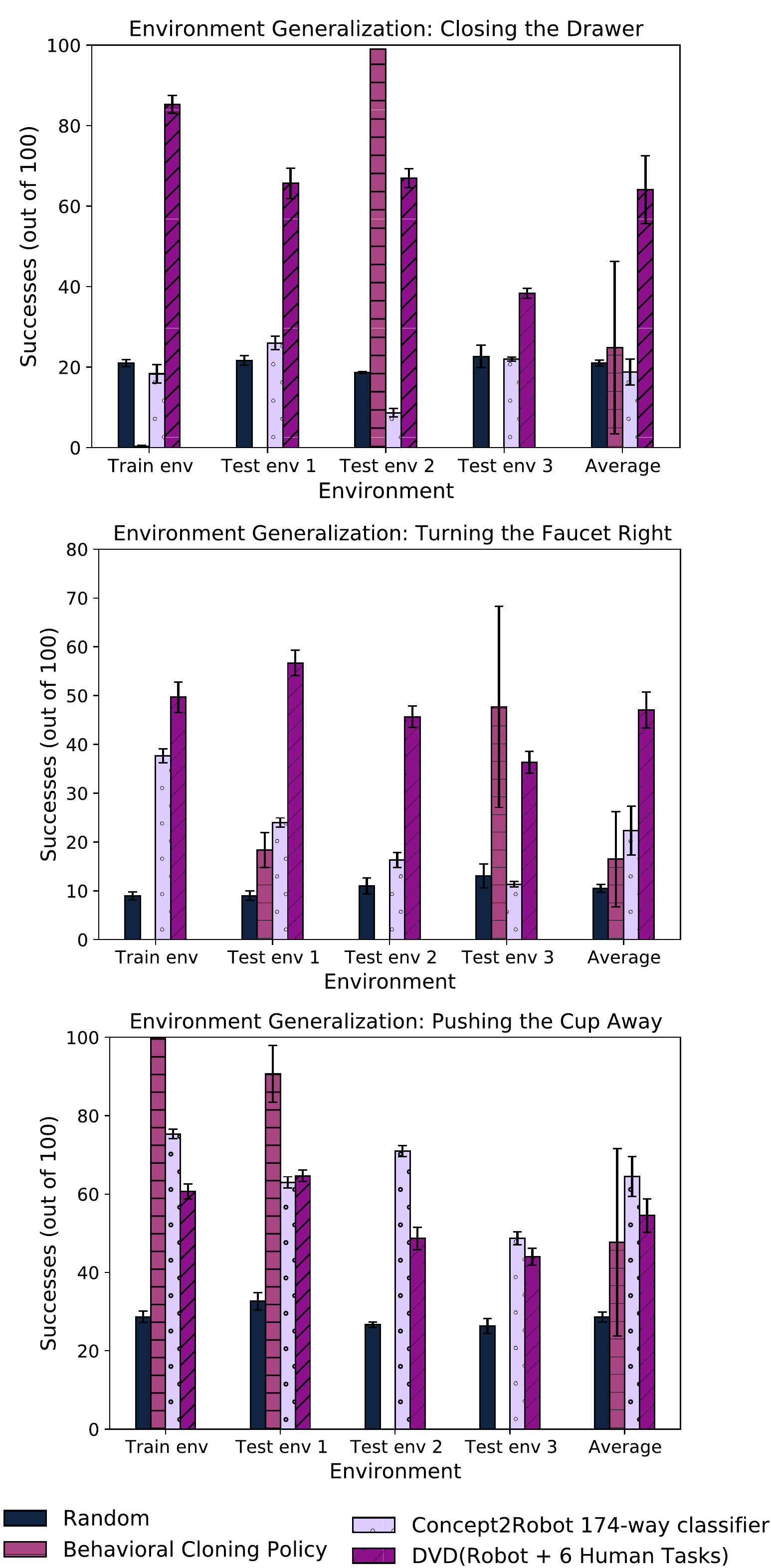}
    \caption{\small \textbf{Environment Generalization Prior Work Comparison.} We compare \algo's performance to Concept2Robot, the most relevant work, a demo-conditioned behavioral cloning policy, and a random policy. On average across environments, \algo performs around or over 30\% better than the next-best performing method on two of the three tasks. Each bar shows the average success rate over all 3 target tasks, computed over 3 seeds of 100 trials, with error bars denoting standard error.}
    \vspace{-0.6cm}
    \label{fig:appendix-env-gen_prior}
\end{figure}

In Figure~\ref{fig:appendix-env-gen_prior}, we present results on the individual tasks across all four environments for \algo trained with 6 tasks worth of human videos compared with our three comparisons: Concept2Robot~\cite{shao2020concept}, a demo-conditioned behavior cloning policy, and a random policy. On average over all of the environments, \algo performs over 40\% better on the drawer task and 30\% better on the faucet task than the next best performing method. It also performs reasonably on the cup task; Concept2Robot just performs particularly well on that task since it often chooses to push the cup away no matter which task is specified. The behavioral cloning policy has somewhat erratic behavior, mimicking the trajectory for one of the target tasks in each environment and doing well on that task but not on the other tasks. Hence, we see that both Concept2Robot and the behavioral cloning policy are not able to provide effective \emph{multi-task} reward signals for each environment. 

\begin{figure*}%
    \centering
    \vspace{0.1cm}
    \includegraphics[width=0.9\linewidth]{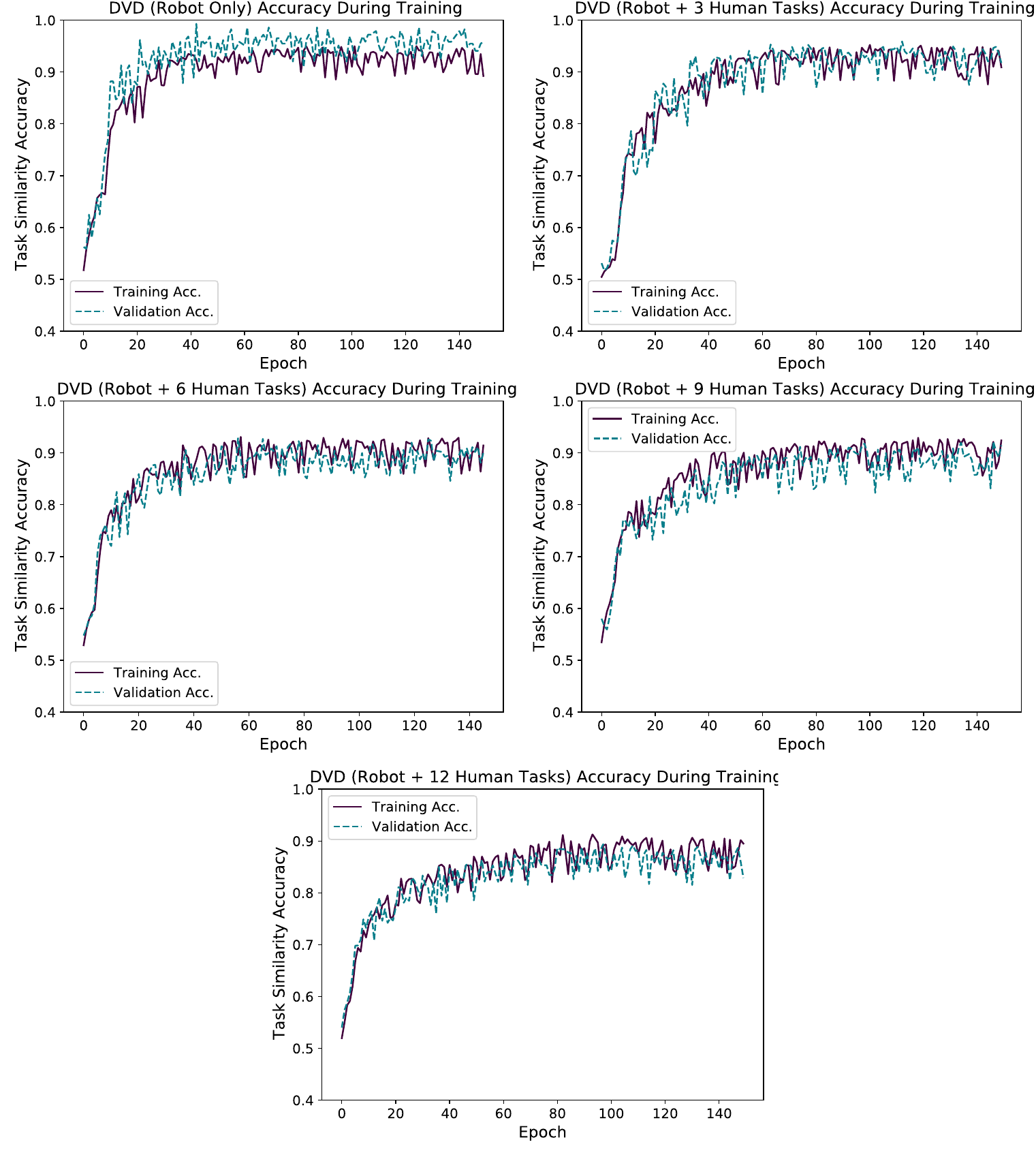}
    \caption{\small \textbf{Accuracy Curves During \algo Training.} We plot both training and validation accuracies over the course of training \algo for 150 epochs with varying amounts of human data. The accuracies gradually decrease as more human videos are added, but we find that this trade-off is worthwhile for greater generalization capabilities.}
    \vspace{-0.6cm}
    \label{fig:appendix-accuracies}
\end{figure*}

Additionally, in Figure~\ref{fig:appendix-accuracies}, we show the accuracy curves on the training and validation sets while training \algo. We see unsurprisingly that the model trained only on three tasks of robot demonstrations (Robot Only) has the highest validation accuracy at 99\%. However, while adding human videos significantly increases the difficulty of the optimization, the models remain generally robust, with \algo trained on robot data and 12 tasks worth of human videos still obtaining 89\% validation accuracy. We find in our experiments in Section~\ref{sec:exps} that this trade-off in discriminator accuracy from adding human videos to the training set results in much greater ability to generalize to unseen environments and tasks. 

\begin{figure*}[t]%
    \centering
    \vspace{0.1cm}
    \includegraphics[width=0.99\linewidth]{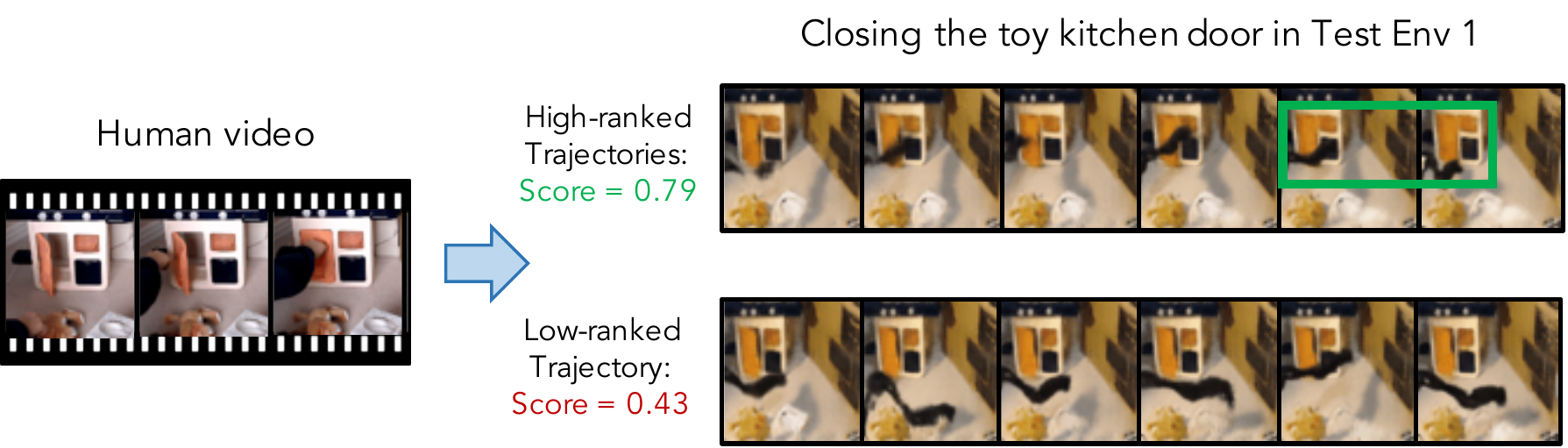}
    \caption{\small \textbf{Rankings on the real robot.} Examples of predicted trajectories on the WidowX200 that are ranked high and low for the task of closing an unseen toy kitchen door. \algo gives the predicted trajectory where the door is closed a high similarity score and the predicted trajectory where the door stays open a low similarity score.}
    \label{fig:realrobot_rankings}
\end{figure*}

Finally, in Figure~\ref{fig:realrobot_rankings}, we include examples on the real Widowx200 of predicted trajectories and their similarity scores with a human video demonstration given by \algo. We see that \algo highly ranks trajectories that are completing the same task as demonstrated in the given human video.

\end{document}